\documentclass[letterpaper]{article} 
\usepackage{aaai2026}  
\usepackage{times}  
\usepackage{helvet}  
\usepackage{courier}  
\usepackage[hyphens]{url}  
\usepackage{graphicx} 
\def\UrlFont{\rm}  
\usepackage{natbib}  
\usepackage{caption} 
\usepackage{algorithm}
\usepackage{algorithmic}
\usepackage{amsmath}

\usepackage{booktabs} 
\usepackage{amsfonts}
\usepackage{tabularx} 
\usepackage{array}    
\usepackage{xcolor}         
\usepackage{multirow}       
\renewcommand{\arraystretch}{1.2} 

\usepackage{siunitx}        

\newcommand{\datastd}[2]{#1 {$\pm #2$}}

\usepackage{newfloat}
\usepackage{listings}
\DeclareCaptionStyle{ruled}{labelfont=normalfont,labelsep=colon,strut=off} 
\floatstyle{ruled}
\newfloat{listing}{tb}{lst}{}
\floatname{listing}{Listing}
\title{SPIRAL: Symbolic LLM Planning via Grounded and Reflective Search}
\author{
    Yifan Zhang\textsuperscript{\rm 1,2}\thanks{Work done during an AI research internship at IBM T.J. Watson Research Center.}, 
    Giridhar Ganapavarapu\textsuperscript{\rm 2}, 
    Srideepika Jayaraman\textsuperscript{\rm 2}, \\
    Bhavna Agrawal\textsuperscript{\rm 2}, 
    Dhaval Patel\textsuperscript{\rm 2}, 
    Achille Fokoue\textsuperscript{\rm 2} 
}
\affiliations{
    \textsuperscript{\rm 1}Vanderbilt University, Nashville, TN, USA\\
    \textsuperscript{\rm 2}IBM T.J. Watson Research Center, Yorktown Heights, NY, USA\\

    \{yifan.zhang.2\}@vanderbilt.edu, \{giridhar.ganapavarapu, j.srideepika\}@ibm.com\\
    \{bhavna, pateldha, achille\}@us.ibm.com
}

\begin{document}

\makeatletter
\let\orig@fnsymbol\@fnsymbol
\def\@fnsymbol#1{%
  \ensuremath{%
    \ifcase#1 \or \dagger \or \ddagger \or \S \or \P \or \| %
    \else \orig@fnsymbol{#1}\fi}}
\makeatother

\maketitle


\begin{abstract}
Large Language Models (LLMs) often falter at complex planning tasks that require exploration and self-correction, as their linear reasoning process struggles to recover from early mistakes. While search algorithms like Monte Carlo Tree Search (MCTS) can explore alternatives, they are often ineffective when guided by sparse rewards and fail to leverage the rich semantic capabilities of LLMs. We introduce SPIRAL (Symbolic LLM Planning via Grounded and Reflective Search), a novel framework that embeds a cognitive architecture of three specialized LLM agents into an MCTS loop. SPIRAL’s key contribution is its integrated planning pipeline where a Planner proposes creative next steps, a Simulator grounds the search by predicting realistic outcomes, and a Critic provides dense reward signals through reflection. This synergy transforms MCTS from a brute-force search into a guided, self-correcting reasoning process. On the DailyLifeAPIs and HuggingFace datasets, SPIRAL consistently outperforms the default Chain-of-Thought planning method and other state-of-the-art agents. More importantly, it substantially surpasses other state-of-the-art agents; for example, SPIRAL achieves 83.6\% overall accuracy on DailyLifeAPIs, an improvement of over 16 percentage points against the next-best search framework, while also demonstrating superior token efficiency. Our work demonstrates that structuring LLM reasoning as a guided, reflective, and grounded search process yields more robust and efficient autonomous planners. The source code, full appendices, and all experimental data are available for reproducibility at the official project repository.
\end{abstract}


\begin{links}
    \link{Code}{https://github.com/IBM/SPIRAL}
\end{links}

\section{Introduction}
\label{sec:introduction}

The recent evolution of Large Language Models (LLMs) has marked a paradigm shift from pure text generation towards the development of autonomous agents capable of complex, goal-oriented behavior \cite{zhao2023}. A key capability of these agents is their ability to formulate symbolic plans and interact with external tools to accomplish tasks \cite{schick2023toolformer}. Foundational approaches for this planning process often rely on prompting the LLM to interleave reasoning "thoughts" with "actions" in a linear sequence \cite{wei2022, yao2023react}. However, the reliability of this auto-regressive generation is a critical bottleneck; the plans produced are often brittle, where a single logical error can derail the entire process without a mechanism for structured deliberation or backtracking. Consequently, enhancing the robustness of LLM agents necessitates a move beyond simple linear prompting towards more sophisticated planning paradigms.

To address this challenge, integrating tree search frameworks has shown potential for adding structured exploration to the planning process \cite{yao2023tree, zhou2024language}. While MCTS \cite{silver2017mastering} is a particularly compelling choice, applying it effectively presents significant challenges. Standard implementations often treat the LLM as a black-box policy with sparse rewards, leading to inefficient exploration \cite{zhang2024rest}. Furthermore, without a grounding world model, agents may explore syntactically valid but practically nonsensical paths \cite{hao2023reasoning}. This highlights a critical gap for a framework that synergizes MCTS's exploratory power with LLM semantics, guided by dense, semantic-aware feedback.

\begin{figure}[t]
    \centering
    \includegraphics[width=0.95\columnwidth]{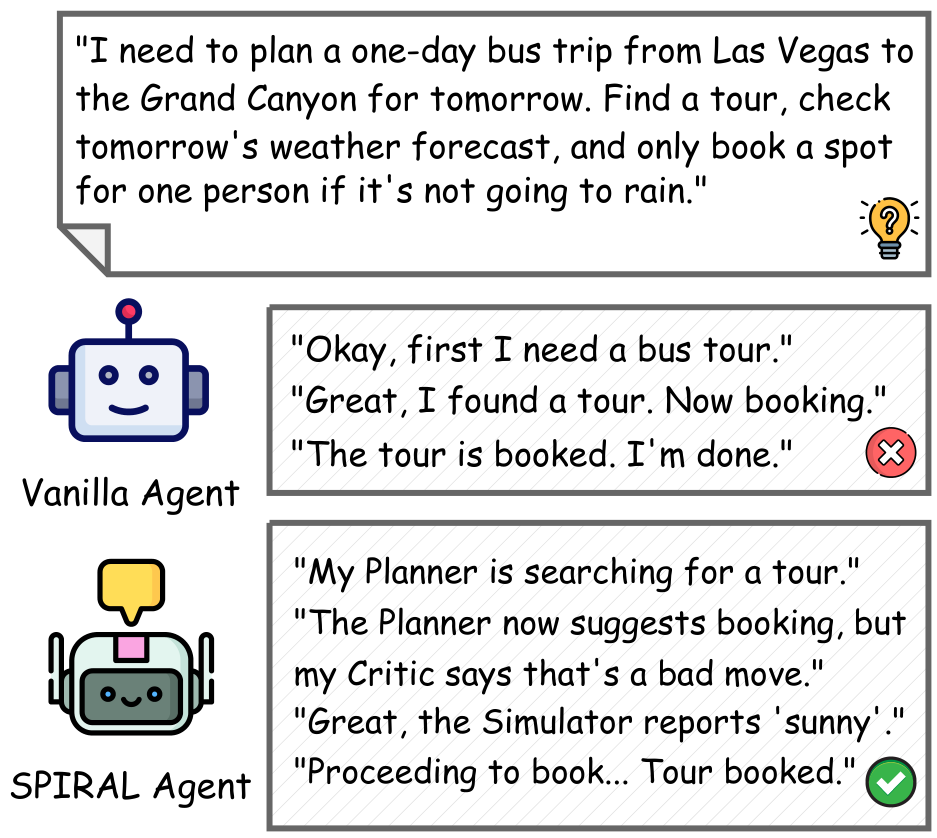}
    \caption{A comparison of a Vanilla Agent and the SPIRAL Agent on a conditional planning task. The Vanilla Agent follows a linear but flawed plan, while SPIRAL uses its grounded and reflective search to correctly handle the weather-based constraint before acting.}
    \label{fig:spiral_vs_vanilla}
\end{figure}

To overcome these limitations, we propose \textbf{SPIRAL} (Symbolic LLM Planning via Grounded and Reflective Search), a framework designed to address the strategic failures of linear planners. The motivation for our approach is illustrated in Figure~\ref{fig:spiral_vs_vanilla}. Faced with a simple conditional planning task, a standard "vanilla" agent follows a plausible but incorrect plan, failing to respect the user's constraints. In contrast, SPIRAL successfully navigates the task by employing a more deliberative process. It reframes LLM-driven MCTS by introducing a cognitive architecture of three specialized LLM agents that work in concert: a \textit{Planner} to propose actions, a \textit{Critic} to reflect on their strategic soundness, and a \textit{Simulator} to ground the plan in plausible outcomes. The synergy between these agents transforms the search process, allowing SPIRAL to intelligently prune flawed reasoning paths and converge on robust solutions.

SPIRAL is built on the key insight that LLM-based planning can be made more robust by structuring it as a grounded and reflective search process. We achieve this by embedding three core principles within the MMCTS framework: 1) Decomposition, treating planning as a cognitive task fulfilled by specialized agentic roles; 2) Grounding, where a learned world model grounds exploration in plausible consequences; and 3) Reflection, using a dense reward signal that evaluates strategic merit to replace sparse terminal rewards. To our knowledge, SPIRAL is the first framework to holistically integrate these principles within a formal search algorithm for LLM agents. Our approach substantially outperforms other state-of-the-art agents; on the DailyLifeAPIs of TaskBench, SPIRAL achieves 83.6\% accuracy, surpassing the next-best search framework by over 16 percentage points.


Our main contributions are fourfold: 1) We propose \textbf{SPIRAL}, a novel framework that integrates a synergistic tri-agent cognitive architecture into Monte Carlo Tree Search to enable structured and dynamic planning. 2) We introduce its core mechanisms: a \textit{Planner} to generate actions, a \textit{Critic} to provide dense reflective feedback, and a \textit{Simulator} to ground the search. Together, these components address the common MCTS challenges of sparse rewards and ungrounded exploration. 3) We evaluate SPIRAL on complex tool-use benchmarks, showing that it substantially outperforms both default planners and state-of-the-art frameworks like ReAct and LATS. 4) Finally, we demonstrate SPIRAL's superior resource efficiency and validate our architecture through ablation studies that confirm the critical role of each component.

\section{Preliminaries}
\label{sec:preliminaries}

SPIRAL's core idea is to frame the complex task of LLM-based planning as a search process that is both grounded and reflective. To achieve this, the paper defines tool-use planning as a sequential decision-making problem within a Markov Decision Process (MDP) framework. The approach then uses the Monte Carlo Tree Search (MCTS) algorithm as its foundational search method.

\subsection{LLM-based Planning as a Search Problem}
We model the task of generating a multi-step tool-use plan as a search problem through a state space, formally defined as a sequential decision-making process by the tuple $(\mathcal{S}, \mathcal{A}, T, R)$. The \textit{state space} $\mathcal{S}$ consists of states $s_t \in \mathcal{S}$, where each state is the history of actions and observations generated so far, often called a chain: $s_t = (a_0, o_1, \dots, a_{t-1}, o_t)$. The \textit{action space} $\mathcal{A}$ is the union of available tool calls, $\mathcal{A}_{\text{tool}}$, and a terminal action, $\mathcal{A}_{\text{term}} = \{\texttt{finish}\}$. The \textit{transition function} $T: \mathcal{S} \times \mathcal{A} \rightarrow \mathcal{S}$ maps a state-action pair to a new state. Given a state $s_t$ and an action $a_t$ from a policy $\pi(a_t|s_t)$, we approximate this transition with a world model, $\mathcal{W}(o_{t+1}|s_t, a_t)$, which generates a plausible observation $o_{t+1}$. The subsequent state is then $s_{t+1} = s_t \oplus (a_t, o_{t+1})$. Finally, the \textit{reward function} $R: \mathcal{S} \times \mathcal{A} \rightarrow \mathbb{R}$ provides feedback. A key challenge in this domain is that the reward signal is sparse; it is zero for all intermediate steps and non-zero only upon termination, where $R(s_T, a_T)$ evaluates the final plan's success.

The overall objective is to find an optimal policy, $\pi^*$, which generates a sequence of actions $(a_0, a_1, \dots, a_T)$ that maximizes the expected cumulative reward:
\begin{equation}
    \pi^* = \arg\max_{\pi} \mathbb{E}\left[\sum_{t=0}^{T} R(s_t, a_t)\right]
\end{equation}

\subsection{Monte Carlo Tree Search}
MCTS is a heuristic search algorithm particularly well-suited for navigating the large decision spaces inherent in planning problems. The algorithm iteratively builds a search tree to balance the exploration of new paths with the exploitation of known promising ones. Each node $n$ in the tree represents a state $s_n$ and stores two key statistics: its total accumulated reward, or \textit{value} $v_n$, and its \textit{visit count} $c_n$.

A canonical MCTS iteration consists of four steps. The process begins with a \textit{selection} phase, where the algorithm traverses the tree from the root by recursively selecting the child node $i$ of a parent node $p$ that maximizes the Upper Confidence Bound for Trees (UCT) score:
\begin{equation}
    \text{UCT}_i = \frac{v_i}{c_i} + C \sqrt{\frac{\ln(c_p)}{c_i}}
\end{equation}
where $C$ is an exploration constant. Once a leaf node is reached, the \textit{expansion} phase adds a new child node by taking a valid, unexplored action. From this new node, a \textit{simulation} (or rollout) is performed, typically using random actions until a terminal state is reached, to produce an estimated value $\hat{v}$. Finally, in the \textit{backpropagation} phase, this value is propagated up the tree to the root. For each ancestor node $n_i$ on the path, the statistics are updated as $c_i \leftarrow c_i + 1$ and $v_i \leftarrow v_i + \hat{v}$. This update process ensures that the knowledge gained from the rollout is distributed to all relevant parent nodes in the search path. Over many iterations, this allows the UCT selection policy to become more accurate, effectively guiding the search towards the most promising regions of the decision space. Our work adapts this traditional MCTS framework by replacing its core components with specialized LLM agents, as detailed in the next section.

\begin{figure*}[t]
\centering
\includegraphics[width=1\textwidth]{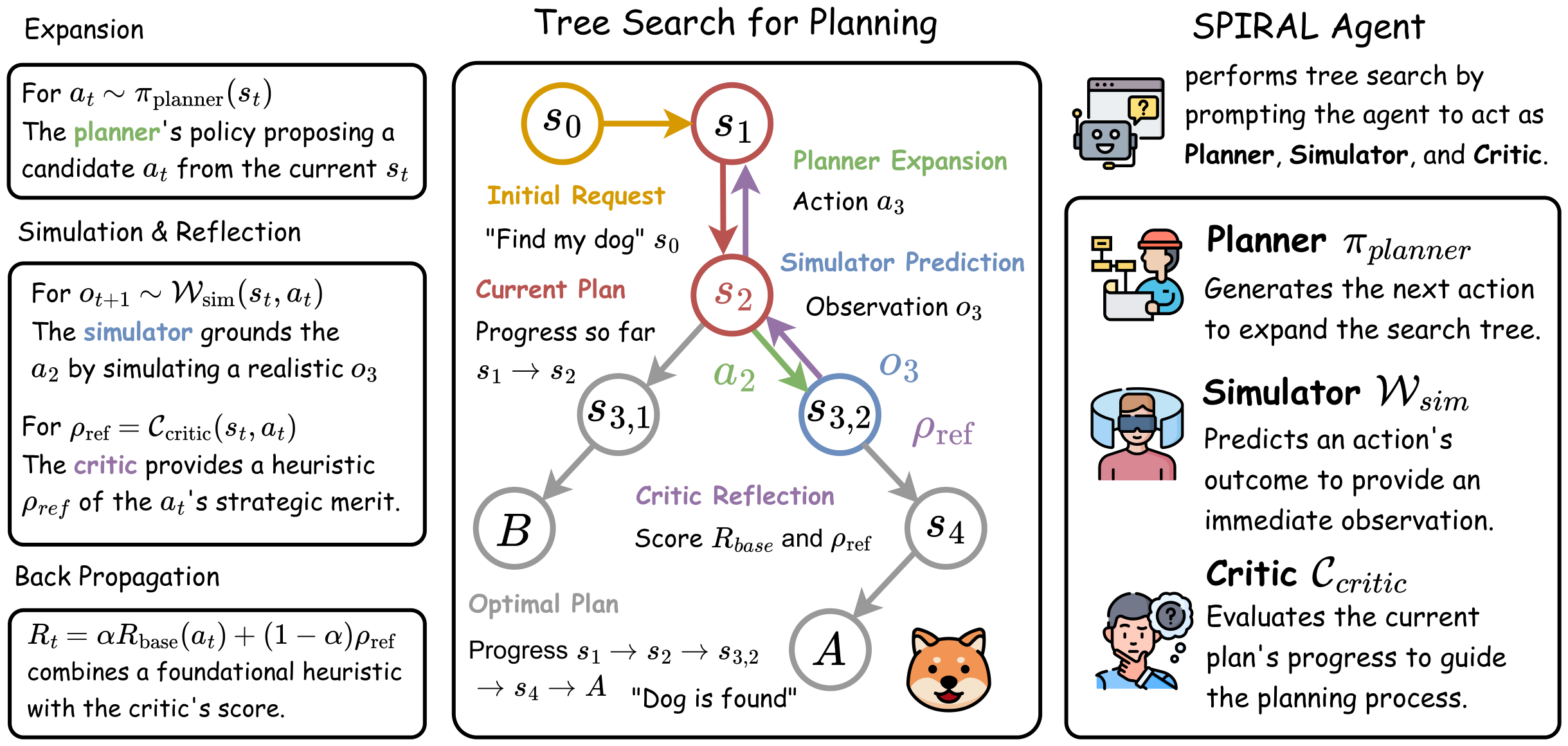} 
\caption{An overview of the SPIRAL framework, where a tri-LLM symbolic architecture drives the MCTS loop. \textbf{(1) Expansion:} The Planner proposes a new action ($a_2$) to expand the search from the current state ($s_2$). \textbf{(2) Simulation \& Reflection:} The Simulator provides a grounded observation ($o_3$) for the action, while the Critic generates a strategic reflection score ($\rho_{\text{ref}}$) evaluating the action's merit. \textbf{(3) Backpropagation:} A composite reward, calculated using the Critic's score, is then propagated up the tree to update node values and intelligently guide future selections.}
\label{fig:spiral_overview}
\end{figure*}

\section{The SPIRAL Framework}
\label{sec:methodology}


To overcome the limitations of sparse rewards and unguided exploration, we introduce the SPIRAL framework (Figure~\ref{fig:spiral_overview}). Our approach integrates a multi-agent cognitive architecture into MCTS, decomposing planning into specialized roles for proposing, grounding, and evaluating actions. We detail this architecture, the reward shaping mechanism, and the search algorithm below.

\subsection{SPIRAL's Cognitive Architecture}
\label{subsec:architecture}

SPIRAL's design decomposes planning into specialized cognitive roles, efficiently instantiating a \textit{Planner}, \textit{Simulator}, and \textit{Critic} within a single LLM (Figure~\ref{fig:spiral_overview}). This creates a robust, self-correcting process that leverages in-context learning, allowing SPIRAL to adapt to new problems without fine-tuning.

\subsection{Agent Roles and Functionality}
\label{subsec:agent_roles}
Each agent in SPIRAL's cognitive architecture is responsible for a specific function within the MCTS search, corresponding to the roles and prompts outlined in Figure 2.

\textbf{The Planner ($\pi_{\text{planner}}$).} The Planner drives the \textbf{Expansion} phase of the search and acts as the creative strategist. Given the current problem description and the action-observation history, the Planner's function is to generate the next candidate action to expand the search tree. As the exploration policy of the framework, it is designed to propose diverse and contextually relevant next steps, providing the raw material for the search process.

\textbf{The Simulator ($\mathcal{W}_{\text{sim}}$).} The Simulator functions as a learned world model to ground the search process during the \textbf{Simulation \& Reflection} phase. When the Planner proposes an action, the Simulator's role is to predict a plausible, natural language observation that would result from executing that action. This grounds the search in realistic consequences, allowing the agent to plan based on the likely outcomes of its actions rather than operating in a vacuum of pure reason.


\textbf{The Critic ($\mathcal{C}_{\text{critic}}$).} The Critic serves as the logical evaluator during the \textbf{Simulation \& Reflection} phase, providing dense, strategic feedback. It assesses the strategic merit of the Planner's action, producing a quantitative reflection score, $\rho_{\text{ref}}$. This score is combined with a foundational heuristic, $R_{\text{base}}$, to form a composite reward, $R_t$. This reward is used during backpropagation to guide the search toward strategically sound plans, directly addressing the MCTS challenge of sparse rewards. This use of a dense, semantic score for backpropagation aligns with recent work on using semantic feedback to guide reasoning, e.g., TextGrad and CodeGrad~\cite{yuksekgonul2024textgrad,zhang2025codegrad}.


\subsection{Reflection-Driven Reward Shaping}
\label{subsec:reward_shaping}
To address sparse rewards, SPIRAL introduces Reflection-Driven Reward Shaping, generating a dense, semantic-aware signal $R_t$ for use during \textbf{Backpropagation}. Unlike outcome-based rewards, $R_t$ is calculated immediately for each expanded node as $R_t = \alpha R_{\text{base}}(a_t) + (1-\alpha) \rho_{\text{ref}}$, blending a validity heuristic $R_{\text{base}}$ with the Critic's strategic score $\rho_{\text{ref}}$ via hyperparameter $\alpha \in [0, 1]$. Crucially, this formulation converts qualitative semantic evaluations into a numerical scalar, allowing the algorithm to directly apply standard MCTS backpropagation updates to the value ($v$) and visit count ($c$) statistics of ancestor nodes. This guides the search toward paths that are both valid and strategically sound, significantly improving efficiency.

\begin{table}[t]
  \centering
  \resizebox{1\columnwidth}{!}{%
  \begin{tabular}{lccccc}
    \toprule
    \textbf{Approach} & \textbf{TU} & \textbf{ME} & \textbf{SF} & \textbf{SR} & \textbf{DU} \\
    \midrule
    CoT \citep{wei2022} & $\times$ & \checkmark & $\times$ & $\times$ & $\times$ \\
    ReAct \citep{yao2023react}    & \checkmark & $\times$ & \checkmark & $\times$ & $\times$ \\
    Reflexion \citep{shinn2023reflexion} & \checkmark & $\times$ & \checkmark & $\times$ & $\times$ \\
    MCTS \citep{swiechowski2023monte} & \checkmark & \checkmark & $\times$ & $\times$ & $\times$ \\
    ToT \citep{yao2023tree}        & $\times$ & \checkmark & $\times$ & \checkmark & \checkmark \\
    RAFA \citep{liu2023reason}     & \checkmark & \checkmark & \checkmark & \checkmark & $\times$ \\
    LATS \citep{zhou2024language}& \checkmark & \checkmark & \checkmark & \checkmark & $\times$ \\
    \midrule
    \textbf{SPIRAL (Ours)}        & \checkmark & \checkmark & \checkmark & \checkmark & \checkmark \\
    \bottomrule
  \end{tabular}%
  } 
  \caption{Comparison of agent frameworks. Columns: Tool Use~(TU), Multi-path Exploration~(ME), Step-wise Feedback~(SF), Strategic Reflection~(SR), and Dynamic Policy Update~(DU). SPIRAL unifies these capabilities.}
  \label{tab:related_work_comparison}
\end{table}

\subsection{Grounded and Reflective Tree Search}
\label{subsec:search_algorithm}
The SPIRAL framework culminates in the Grounded and Reflective Tree Search algorithm, which integrates the cognitive architecture and the reward shaping mechanism into the four canonical stages of MCTS. The full process for each iteration is detailed below.

\textbf{Selection.} The algorithm begins by traversing the existing search tree from the root. At each level, it recursively selects the child node that maximizes the Upper Confidence Bound for Trees (UCT) score (Equation 2), balancing exploitation of known high-value paths with exploration of less-visited ones. This process continues until a leaf node $n_L$ is reached, identifying a promising frontier for expanding the search.


\textbf{Expansion.} Upon reaching a leaf node $n_L$, the \textit{Planner} ($\pi_{\text{planner}}$) leverages the accumulated interaction history to generate a single candidate action $a_L$. This proposal creates a new child node, extending the search frontier with a diverse and contextually relevant step that logically progresses the plan.


\textbf{Simulation \& Reflection.} This stage replaces the expensive, noisy random rollout of MCTS with a deterministic one-step lookahead that is grounded and semantically evaluated. First, the \textit{Simulator} ($\mathcal{W}_{\text{sim}}$) grounds the action $a_L$ by generating a realistic observation $o_{L+1}$ to form a new state $s_{L+1}$. Concurrently, the \textit{Critic} ($\mathcal{C}_{\text{critic}}$) performs reflection, evaluating the action's strategic merit to produce the reflective score $\rho_{\text{ref}}$.


\textbf{Backpropagation.} This final phase uses Reflection-Driven Reward Shaping to calculate a composite reward ($R_t$) for the new node using the semantic formulation defined earlier. This reward is then backpropagated up the path to the root, updating the value ($v$) and visit count ($c$) of each ancestor. Propagating a reward based on strategic merit, rather than a noisy outcome, makes the tree's stored values more reliable indicators of a plan's intrinsic quality, leading to a more intelligent and efficient search.

As summarized in Table~\ref{tab:related_work_comparison}, SPIRAL synthesizes capabilities prior frameworks only partially address, integrating multi-path exploration with tool use. It uniquely enriches this search with two forms of dense, internal feedback: step-wise grounding from the Simulator and strategic guidance from the Critic. The backpropagation of this reflective feedback enables a dynamic policy update at each step, yielding a more robust, deliberative, and adaptive planner. The complete pseudocode is available in Appendix A.


\begin{table*}[t]
\centering
\resizebox{1\textwidth}{!}{%
\begin{tabular}{@{}llcccccc@{}}
\toprule
& & \multicolumn{3}{c}{\textbf{DailyLifeAPIs}} & \multicolumn{3}{c}{\textbf{HuggingFace}} \\
\cmidrule(lr){3-5} \cmidrule(lr){6-8}
\textbf{Model} & \textbf{Method} & \textbf{Simple Acc. (\%)} & \textbf{Complex Acc. (\%)} & \textbf{Overall Acc. (\%)} & \textbf{Simple Acc. (\%)} & \textbf{Complex Acc. (\%)} & \textbf{Overall Acc. (\%)} \\
\midrule

DeepSeek-V2.5 & CoT (k=1) & \datastd{85.12}{6.42} & \datastd{59.22}{5.86} & \datastd{66.60}{4.90} & \datastd{93.84}{0.97} & \datastd{67.92}{1.97} & \datastd{75.77}{1.57} \\
              & CoT (k=3) & \datastd{86.04}{4.09} & \datastd{60.91}{3.83} & \datastd{67.90}{3.87} & \datastd{94.98}{1.35} & \datastd{71.94}{3.06} & \datastd{79.34}{2.42} \\
              & CoT (k=5) & \datastd{84.81}{2.06} & \datastd{62.41}{5.34} & \datastd{68.83}{4.19} & \datastd{94.44}{1.34} & \datastd{71.16}{1.46} & \datastd{78.61}{1.40} \\
\cmidrule(lr){2-8}
              & \textbf{SPIRAL} & \bfseries \datastd{94.33}{3.95} & \bfseries \datastd{89.89}{3.57} & \bfseries \datastd{91.24}{2.65} & \bfseries \datastd{98.89}{0.91} & \bfseries \datastd{95.81}{1.23} & \bfseries \datastd{96.84}{0.65} \\
\midrule

Llama 3.3 70B & CoT (k=1) & \datastd{94.97}{1.85} & \datastd{94.68}{2.67} & \datastd{94.87}{1.80} & \datastd{93.10}{0.71} & \datastd{92.18}{1.67} & \datastd{92.48}{1.29} \\
              & CoT (k=3) & \datastd{95.26}{1.73} & \datastd{94.82}{1.58} & \datastd{94.88}{1.36} & \datastd{93.85}{1.40} & \datastd{92.29}{1.41} & \datastd{92.79}{1.24} \\
              & CoT (k=5) & \datastd{95.54}{3.63} & \datastd{94.06}{2.08} & \datastd{94.38}{1.88} & \datastd{93.77}{0.97} & \datastd{93.75}{0.67} & \datastd{93.75}{0.68} \\
\cmidrule(lr){2-8}
              & \textbf{SPIRAL} & \bfseries \datastd{95.79}{2.44} & \bfseries \datastd{98.82}{0.82} & \bfseries \datastd{98.35}{0.83} & \bfseries \datastd{99.08}{0.90} & \bfseries \datastd{96.76}{1.25} & \bfseries \datastd{97.44}{0.89} \\
\midrule

Llama 4 Maverick 17B & CoT (k=1) & \datastd{84.43}{6.19} & \datastd{47.59}{4.73} & \datastd{57.95}{5.59} & \datastd{91.81}{1.15} & \datastd{67.78}{1.80} & \datastd{76.43}{0.97} \\
              & CoT (k=3) & \datastd{87.72}{5.09} & \datastd{50.21}{5.35} & \datastd{60.60}{4.16} & \datastd{91.67}{2.40} & \datastd{66.24}{1.82} & \datastd{75.65}{1.31} \\
              & CoT (k=5) & \datastd{89.21}{3.95} & \datastd{48.86}{6.53} & \datastd{60.00}{5.31} & \datastd{90.88}{2.17} & \datastd{68.34}{0.83} & \datastd{77.09}{0.76} \\
\cmidrule(lr){2-8}
              & \textbf{SPIRAL} & \bfseries \datastd{93.64}{2.96} & \bfseries \datastd{79.53}{4.44} & \bfseries \datastd{83.31}{4.11} & \bfseries \datastd{93.63}{1.92} & \bfseries \datastd{92.94}{0.56} & \bfseries \datastd{93.04}{0.89} \\
\midrule

Phi 4 14B     & CoT (k=1) & \datastd{89.51}{3.34} & \datastd{83.88}{2.54} & \datastd{85.67}{2.67} & \datastd{95.27}{1.20} & \datastd{90.38}{0.75} & \datastd{92.08}{0.26} \\
              & CoT (k=3) & \datastd{89.81}{4.86} & \datastd{85.19}{4.00} & \datastd{86.24}{3.61} & \datastd{96.86}{1.55} & \datastd{90.72}{1.59} & \datastd{92.84}{0.95} \\
              & CoT (k=5) & \datastd{91.63}{1.65} & \datastd{84.27}{4.38} & \datastd{86.45}{3.44} & \bfseries \datastd{97.49}{1.06} & \datastd{91.72}{2.10} & \datastd{93.71}{1.13} \\
\cmidrule(lr){2-8}
              & \textbf{SPIRAL} & \bfseries \datastd{95.48}{3.63} & \bfseries \datastd{90.53}{2.25} & \bfseries \datastd{91.57}{2.44} & \datastd{96.67}{1.83} & \bfseries \datastd{94.95}{0.82} & \bfseries \datastd{95.48}{0.86} \\
\midrule

Qwen 2.5 72B  & CoT (k=1) & \datastd{90.45}{4.30} & \datastd{89.25}{2.10} & \datastd{89.70}{2.03} & \datastd{90.51}{2.33} & \datastd{85.05}{1.85} & \datastd{86.78}{1.58} \\
              & CoT (k=3) & \datastd{91.01}{3.33} & \datastd{89.02}{1.80} & \datastd{89.64}{1.14} & \datastd{91.45}{3.79} & \datastd{86.57}{1.80} & \datastd{88.16}{1.11} \\
              & CoT (k=5) & \datastd{92.10}{3.39} & \datastd{88.47}{1.21} & \datastd{89.50}{1.37} & \datastd{91.50}{2.61} & \datastd{87.09}{2.00} & \datastd{88.47}{1.65} \\
\cmidrule(lr){2-8}
              & \textbf{SPIRAL} & \bfseries \datastd{94.09}{3.94} & \bfseries \datastd{100.00}{0.00} & \bfseries \datastd{97.69}{1.23} & \bfseries \datastd{97.73}{1.06} & \bfseries \datastd{98.52}{0.49} & \bfseries \datastd{97.08}{0.54} \\

\bottomrule
\end{tabular}
}
\caption{
    Performance comparison of CoT and our proposed SPIRAL method. We report the mean ± standard deviation over 5 runs with fixed seeds. 
    The results for our \textbf{SPIRAL} method are highlighted unless a baseline performs better. 
    \textbf{Simple Acc.}: Simple Task Accuracy (\%), 
    \textbf{Complex Acc.}: Complex Task Accuracy (\%), 
    \textbf{Overall Acc.}: Overall Accuracy (\%). For CoT, $k$ refers to the number of self-consistency levels used. Full hyperparameter details for all methods are listed in Appendix B.
}
\label{tab:main_results_final_polished}
\end{table*}

\section{Experimental Setup}
\label{sec:setup}


We comprehensively evaluate SPIRAL's performance, efficiency, and robustness through three main analyses: comparing against standard Chain-of-Thought (CoT) baselines, benchmarking against state-of-the-art agent frameworks, and conducting a detailed ablation study to validate our design. Collectively, these experiments are designed to isolate the impact of our grounded, reflective search mechanism on both solution quality and computational cost.

\subsection{Datasets and Tasks}
\label{subsec:datasets}
Our evaluation uses two benchmarks for complex, multi-step tool utilization from the TaskBench suite~\cite{shen2024taskbench}: `dailylifeapis` and `huggingface`. To ensure a robust and reproducible evaluation, all experiments are conducted using five fixed random seeds, and we report the \textit{mean and standard deviation} across these runs. Appendix C details the data sampling and pre-processing methodology used in the experiments.

\subsection{Implementation and Evaluation}
\label{subsec:implementation_evaluation}

All agents are implemented using a suite of state-of-the-art models: DeepSeek-V2.5, Llama 3.3 70B, Llama 4 Maverick 17B, Phi 4 14B, and Qwen 2.5 72B. Model interactions were conducted via a dedicated internal research cluster (see Appendix D for details) and orchestrated using the LangChain library. Our SPIRAL agent uses a Planner temperature of 0.1, an MCTS budget of 50 iterations, an exploration constant $C=1.5$, and a reward shaping hyperparameter $\alpha=0.5$. We include a brief sensitivity analysis for key hyperparameters, such as the MCTS budget and the reward shaping coefficient $\alpha$, in Appendix B, which confirms the robustness of our chosen configuration. The baseline frameworks, whose capabilities are compared in Table~\ref{tab:related_work_comparison}, are configured for a strong and fair comparison, with full details available in our experiment scripts.

\textbf{Evaluation Metrics.} Our primary evaluation metric is Success Rate, which measures the percentage of tasks correctly solved. To provide a more nuanced analysis, we report this metric across three categories: Overall Accuracy, Simple Task Accuracy on single-step plans, and Complex Task Accuracy on multi-step plans. To analyze resource costs, we also report two efficiency metrics: Token Efficiency, defined as the success rate per 10,000 tokens consumed, and API Call Efficiency, defined as the success rate per LLM call made during the planning process.

\begin{figure}[t]
    \centering
    \includegraphics[width=\columnwidth]{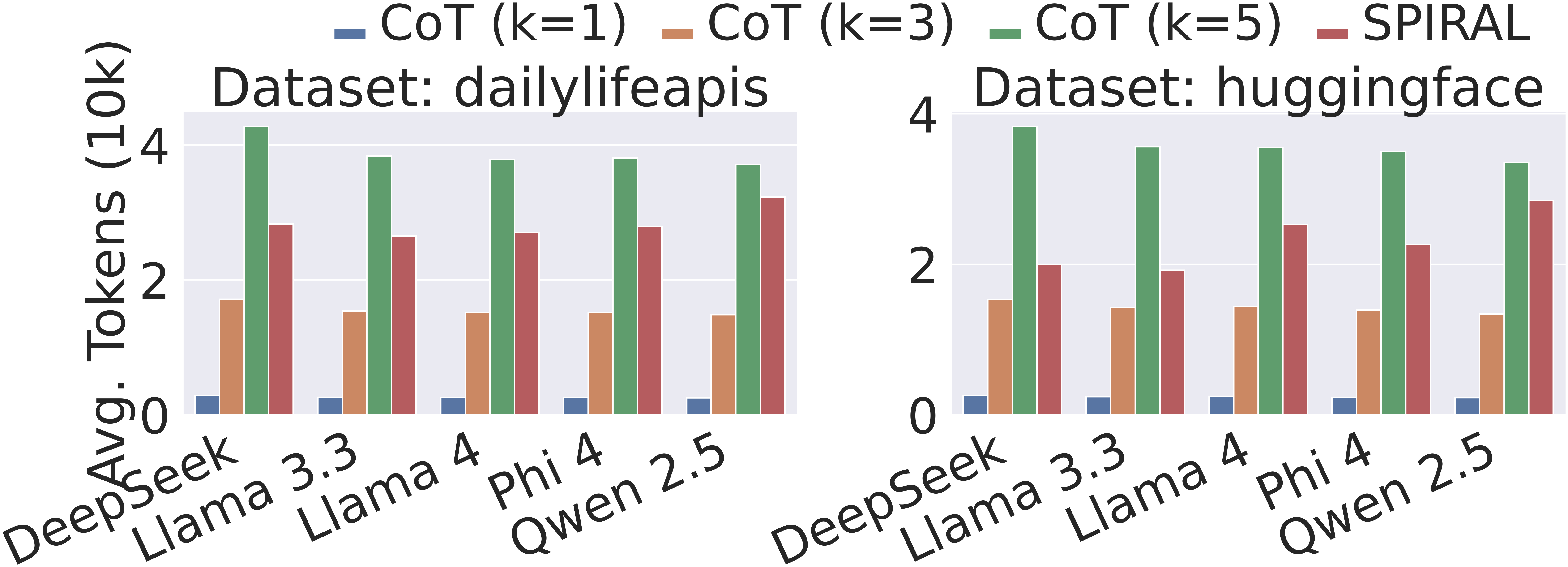}
    \caption{
        Comparison of average token usage per task across different models and methods. 
        Our SPIRAL method consistently reduces token costs compared to CoT baselines on both datasets.
    }
    \label{fig:cost_tokens}
\end{figure}

\begin{figure}[t]
    \centering
    \includegraphics[width=\columnwidth]{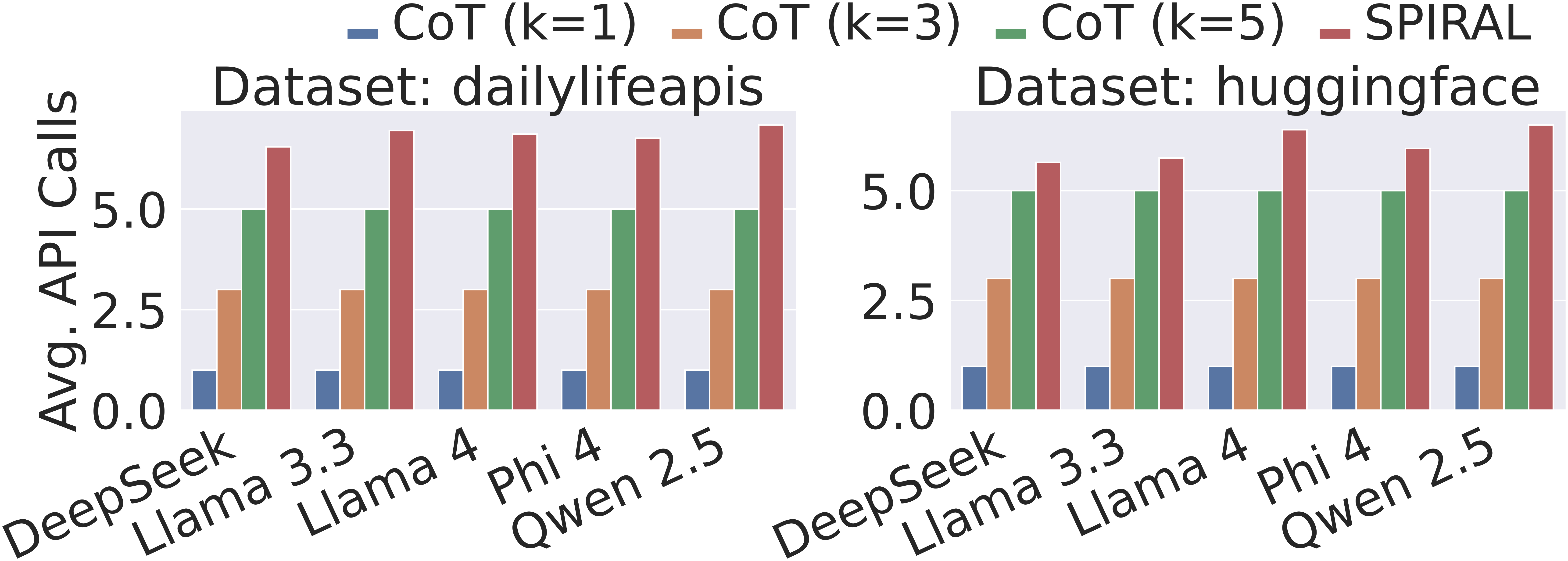}
    \caption{
        Comparison of average API call usage per task across different models and methods. 
        SPIRAL requires more API calls, reflecting its more complex reasoning process.
    }
    \label{fig:cost_api_calls}
\end{figure}

\section{Results and Analysis}
\label{sec:results}


This section empirically evaluates SPIRAL through comparisons against CoT baselines and state-of-the-art frameworks, followed by an ablation study. Together, these analyses confirm SPIRAL's robustness in complex planning and validate the necessity of its cognitive architecture.

\begin{table*}[htbp]
\centering
\resizebox{1\textwidth}{!}{%
\begin{tabular}{@{}llcccccc@{}}
\toprule
& & \multicolumn{3}{c}{\textbf{DailyLifeAPIs}} & \multicolumn{3}{c}{\textbf{HuggingFace}} \\
\cmidrule(lr){3-5} \cmidrule(lr){6-8}
\textbf{Model} & \textbf{Method} & \textbf{Simp. Acc. (\%)} & \textbf{Comp. Acc. (\%)} & \textbf{Overall Acc. (\%)} & \textbf{Simp. Acc. (\%)} & \textbf{Comp. Acc. (\%)} & \textbf{Overall Acc. (\%)} \\
\midrule

Llama 4 Maverick 17B & ReAct \cite{yao2023react} & \datastd{87.79}{6.20} & \datastd{51.38}{6.20} & \datastd{61.60}{6.92} & \datastd{94.81}{0.86} & \datastd{70.95}{1.86} & \datastd{79.54}{1.15} \\
                      & RAFA \cite{liu2023reason} & \datastd{88.90}{6.76} & \datastd{53.83}{4.83} & \datastd{63.75}{6.24} & \datastd{96.32}{0.73} & \datastd{73.75}{1.88} & \datastd{81.86}{1.28} \\
                      & ReAct+RAFA & \datastd{90.19}{4.13} & \datastd{54.94}{4.55} & \datastd{64.90}{5.32} & \datastd{97.47}{0.60} & \datastd{75.56}{1.90} & \datastd{83.43}{1.18} \\
                      & LATS \cite{zhou2024language} & \datastd{91.48}{5.08} & \datastd{57.93}{3.62} & \datastd{67.39}{4.74} & \datastd{98.28}{0.76} & \datastd{78.60}{1.30} & \datastd{85.67}{0.91} \\
\cmidrule(lr){2-8}
                      & \textbf{SPIRAL} & \bfseries \datastd{95.77}{2.94} & \bfseries \datastd{78.79}{3.50} & \bfseries \datastd{83.61}{3.10} & \bfseries \datastd{99.06}{0.89} & \bfseries \datastd{88.38}{8.76} & \bfseries \datastd{92.18}{5.98} \\
\midrule

Phi 4 14B             & ReAct \cite{yao2023react} & \datastd{90.13}{2.26} & \datastd{85.12}{2.64} & \datastd{86.67}{2.58} & \datastd{96.40}{0.67} & \datastd{90.83}{1.07} & \datastd{92.77}{0.77} \\
                      & RAFA \cite{liu2023reason} & \datastd{90.13}{2.26} & \datastd{85.36}{2.63} & \datastd{86.84}{2.60} & \datastd{97.26}{1.43} & \datastd{91.81}{1.09} & \datastd{93.70}{0.87} \\
                      & ReAct+RAFA & \datastd{91.36}{0.66} & \datastd{85.36}{2.63} & \datastd{87.17}{2.20} & \datastd{97.38}{1.44} & \datastd{92.31}{1.11} & \datastd{94.07}{0.73} \\
                      & LATS \cite{zhou2024language} & \datastd{92.61}{2.44} & \datastd{85.82}{2.55} & \datastd{87.84}{2.19} & \datastd{98.74}{0.92} & \datastd{92.94}{1.04} & \datastd{94.96}{0.86} \\
\cmidrule(lr){2-8}
                      & \textbf{SPIRAL} & \bfseries \datastd{97.81}{2.20} & \bfseries \datastd{95.48}{1.98} & \bfseries \datastd{96.17}{1.61} & \bfseries \datastd{99.64}{0.33} & \bfseries \datastd{99.00}{0.26} & \bfseries \datastd{99.23}{0.09} \\
\midrule

Qwen 2.5 72B          & ReAct \cite{yao2023react} & \datastd{94.34}{4.15} & \datastd{94.86}{1.49} & \datastd{94.77}{0.74} & \datastd{98.93}{1.03} & \datastd{96.94}{0.69} & \datastd{97.57}{0.55} \\
                      & RAFA \cite{liu2023reason} & \datastd{96.20}{3.03} & \datastd{95.57}{2.02} & \datastd{95.78}{1.35} & \datastd{99.60}{0.61} & \datastd{97.84}{0.57} & \datastd{98.40}{0.51} \\
                      & ReAct+RAFA & \datastd{96.96}{2.16} & \datastd{96.04}{2.21} & \datastd{96.29}{1.66} & \datastd{99.60}{0.61} & \datastd{98.34}{0.31} & \datastd{98.74}{0.28} \\
                      & LATS \cite{zhou2024language} & \datastd{96.96}{2.16} & \datastd{96.26}{1.88} & \datastd{96.46}{1.51} & \datastd{99.60}{0.61} & \datastd{98.59}{0.56} & \datastd{98.92}{0.31} \\
\cmidrule(lr){2-8}
                      & \textbf{SPIRAL} & \bfseries \datastd{98.79}{1.76} & \bfseries \datastd{100.00}{0.00} & \bfseries \datastd{99.67}{0.46} & \bfseries \datastd{99.86}{0.31} & \bfseries \datastd{99.81}{0.18} & \bfseries \datastd{99.83}{0.18} \\

\bottomrule
\end{tabular}
}
\caption{
    Cascaded accuracy comparison on Llama 4, Phi 4, and Qwen 2.5. Each method was applied to the failures of a CoT (k=1) baseline. 
    We report the mean ± standard deviation over 5 runs with fixed seeds. The best result for each metric is highlighted in \textbf{bold}.
    \textbf{Simp. Acc.}: Simple Task Acc. (\%), 
    \textbf{Comp. Acc.}: Complex Task Acc. (\%), 
    \textbf{Overall Acc.}: Overall Accuracy (\%). Full hyperparameter details for all methods are listed in Appendix B.
}
\label{tab:rq2_sota_comparison}
\end{table*}

\subsection{Baseline Performance Analysis}

We begin by comparing SPIRAL against CoT baselines (Table~\ref{tab:main_results_final_polished}), which SPIRAL consistently outperforms, with the performance gap most pronounced on \textit{complex tasks} requiring backtracking. For instance, on the DailyLifeAPIs benchmark with Llama 4 Maverick 17B, SPIRAL's accuracy on complex tasks (79.53\%) is over 29 percentage points higher than the best CoT variant (50.21\% for k=3). This advantage stems from SPIRAL's MCTS-based exploration, which navigates conditional logic and recovers from mistakes where linear CoT plans fail.

\textbf{Cost-Benefit Analysis.} While superior in accuracy, we also analyze SPIRAL's operational cost (Figure~\ref{fig:cost_tokens} and Figure~\ref{fig:cost_api_calls}). As expected from its deliberative design, SPIRAL uses more API (LLM) calls than CoT methods. However, this increased interaction leads to a more focused search, making it remarkably token-efficient; it consistently consumes fewer total tokens than the more effective CoT baselines (k=3 and k=5). This highlights a key trade-off: SPIRAL's higher number of targeted calls is a more resource-effective strategy than generating fewer, but much longer and often redundant, reasoning paths. While this deliberative search incurs higher wall-clock latency (detailed in Appendix H), our analysis shows that each API call is used with greater strategic purpose, leading to the highest accuracy per interaction (Figure~\ref{fig:cost_api_calls}).

\subsection{Comparison with State-of-the-Art Methods}

\begin{figure}[t]
    \centering
    \includegraphics[width=\columnwidth]{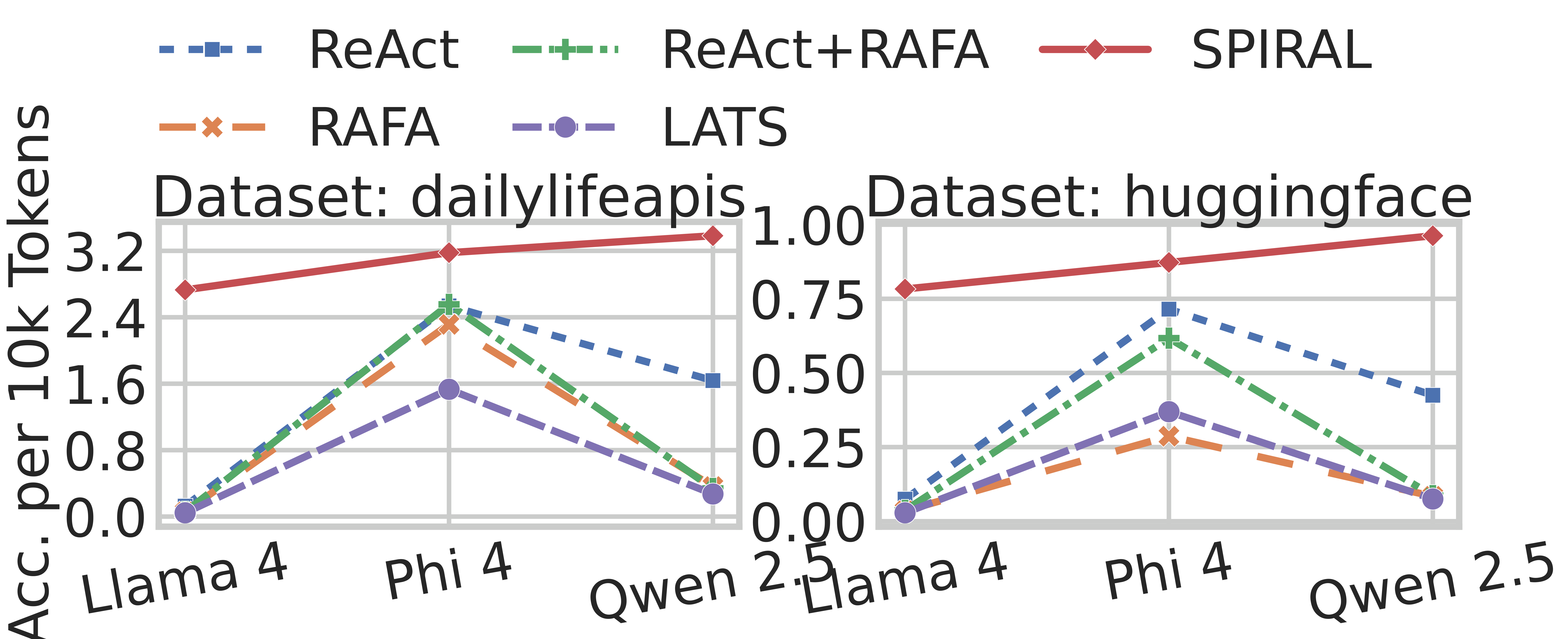}
    \caption{
        Token Efficiency of SOTA methods. This plot shows the final cascaded accuracy achieved per 10,000 tokens consumed. A higher value indicates greater efficiency.
    }
    \label{fig:rq2_token_efficiency}
\end{figure}

\begin{figure}[t]
    \centering
    \includegraphics[width=\columnwidth]{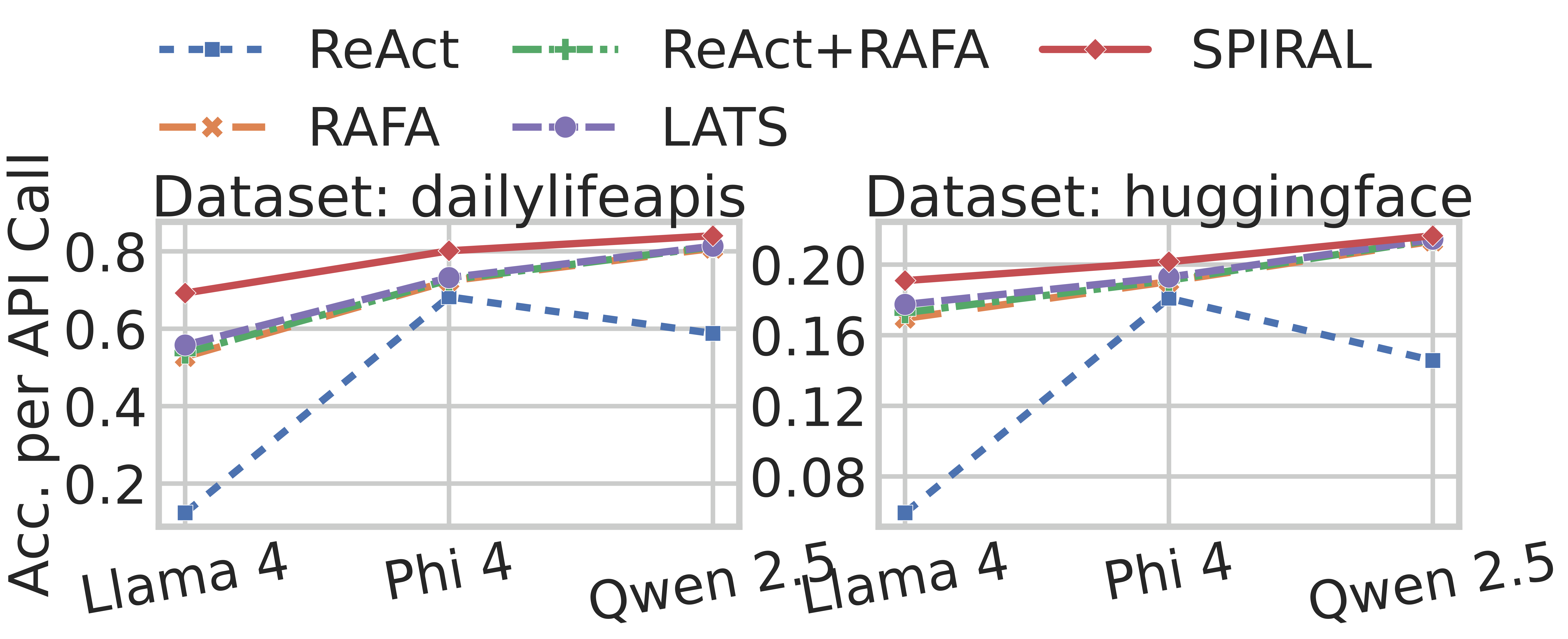}
    \caption{
        API Call Efficiency of SOTA methods. This plot shows the final cascaded accuracy achieved per API call, measuring the effectiveness of each interaction.
    }
    \label{fig:rq2_api_call_efficiency}
\end{figure}

\begin{table}[b]
\centering
\resizebox{0.9\columnwidth}{!}{%
\begin{tabular}{@{}lcc@{}}
\toprule
\textbf{Method} & \textbf{DailyLifeAPIs (\%)} & \textbf{HuggingFace (\%)} \\
\midrule
ToT (S=4, B=3, C=2) & \datastd{60.99}{4.11} & \datastd{82.50}{0.87} \\
ToT (S=5, B=4, C=3) & \datastd{60.99}{4.87} & \datastd{82.54}{0.58} \\
ToT (S=6, B=5, C=4) & \datastd{60.83}{4.12} & \datastd{82.21}{1.48} \\
\midrule
\textbf{SPIRAL}     & \bfseries \datastd{80.83}{3.48} & \bfseries \datastd{88.46}{10.03} \\
\bottomrule
\end{tabular}
}
\caption{
    Cascaded accuracy on Llama 4 Maverick 17B using residual methods. For ToT, configurations are (S=steps, B=breadth, C=candidates). Best results are in \textbf{bold}.
}
\label{tab:rq2_final_accuracy_detailed}
\end{table}

We benchmark SPIRAL against state-of-the-art frameworks, including ReAct, RAFA, LATS, and Tree of Thoughts (ToT), on the challenging set of problems where a baseline CoT agent fails. This cascaded setup rigorously tests each framework's advanced problem-solving capabilities. The results, detailed in Table~\ref{tab:rq2_sota_comparison} and Table~\ref{tab:rq2_final_accuracy_detailed}, demonstrate SPIRAL's superior performance. For instance, on the DailyLifeAPIs dataset using Llama 4 Maverick 17B, SPIRAL achieves an overall cascaded accuracy of 83.61\%, significantly outperforming the next-best method, LATS (67.39\%), and substantially surpassing ToT (~61\%).


\textbf{Resource Efficiency.} Beyond raw accuracy, SPIRAL demonstrates superior resource efficiency, achieving the highest accuracy per token and API call (Figure~\ref{fig:rq2_token_efficiency} and Figure~\ref{fig:rq2_api_call_efficiency}). Although SPIRAL uses more API calls than simple CoT, its guided search ensures each interaction serves a greater strategic purpose than competing frameworks. This confirms that SPIRAL's performance stems from an efficient cognitive architecture rather than brute force.

\subsection{Ablation Study}

\begin{table}[t!]
\centering
\resizebox{0.98\columnwidth}{!}{%
\begin{tabular}{@{}llcc@{}}
\toprule
\textbf{Model} & \textbf{Method} & \textbf{DailyLifeAPIs} & \textbf{HuggingFace} \\
\midrule

\multirow{8}{*}{Llama 3.3 70B} 
& MCTS (N=15)            & \datastd{87.60}{1.31} & \datastd{89.68}{1.43} \\
& MCTS (N=30)            & \datastd{87.11}{1.72} & \datastd{89.12}{0.95} \\
& MCTS (N=50)            & \datastd{86.28}{1.71} & \datastd{89.44}{1.03} \\
\cmidrule(lr){2-4}
& w/o Planner       & \datastd{94.71}{1.90} & \datastd{87.48}{1.42} \\
& w/o Simulator          & \datastd{92.89}{1.11} & \datastd{69.48}{3.02} \\
& w/o Validator          & \datastd{87.27}{2.38} & \datastd{90.08}{1.30} \\
& w/ Uniform Rewards     & \datastd{88.27}{1.79} & \datastd{89.12}{1.06} \\
\cmidrule(lr){2-4}
& \textbf{SPIRAL} & \bfseries \datastd{98.35}{0.83} & \bfseries \datastd{97.44}{0.89} \\
\midrule

\multirow{8}{*}{Llama 4 Maverick 17B} 
& MCTS (N=15)            & \datastd{79.84}{4.66} & \datastd{73.80}{1.39} \\
& MCTS (N=30)            & \datastd{79.83}{4.77} & \datastd{73.64}{1.51} \\
& MCTS (N=50)            & \datastd{80.16}{4.21} & \datastd{73.44}{2.85} \\
\cmidrule(lr){2-4}
& w/o Planner       & \datastd{81.82}{2.41} & \datastd{91.08}{1.46} \\
& w/o Simulator          & \datastd{80.99}{4.89} & \datastd{55.28}{2.87} \\
& w/o Validator          & \datastd{80.00}{4.31} & \datastd{75.16}{2.46} \\
& w/ Uniform Rewards     & \datastd{79.01}{5.31} & \datastd{74.44}{2.06} \\
\cmidrule(lr){2-4}
& \textbf{SPIRAL} & \bfseries \datastd{83.30}{4.11} & \bfseries \datastd{93.04}{0.89} \\

\bottomrule
\end{tabular}
}
\caption{
    Ablation study of SPIRAL components on Llama 3.3 70B and Llama 4. For MCTS baselines, `N`=search iterations. Best results are in \textbf{bold}. Full hyperparameter details for all methods are listed in Appendix B.
}
\label{tab:rq3_ablation_study_final}
\end{table}

To validate our design, we conducted an ablation study, summarized in Table~\ref{tab:rq3_ablation_study_final}. The study confirms each component of the cognitive architecture is critical, with performance degrading most when the Simulator and Critic are removed. Disabling the Simulator (`w/o Simulator`) causes a catastrophic drop in accuracy (e.g., from 97.44\% to 69.48\% on HuggingFace), underscoring the necessity of a grounded world model. Similarly, disabling the Critic's dense feedback (`w/ Uniform Rewards`) significantly impairs performance, validating our reflection-driven approach. Removing the Planner and Validator also results in a noticeable performance decrease. Crucially, the SPIRAL framework outperforms a well-budgeted standard MCTS baseline, confirming the synergy between agents is the primary driver of success, as illustrated by the case studies and failure analysis in Appendix G.

\section{Related Works}
\label{sec:related_works}
Our work is situated at the confluence of research in LLM reasoning and autonomous agent architectures. The rapid growth of this field is captured in several comprehensive surveys on LLMs and autonomous agents \cite{zhao2023, wang2024survey}.

\subsection{Reasoning and Planning with Search}
Recent literature has focused on evolving LLM reasoning from simple linear generation to more robust, structured problem-solving. Initial breakthroughs in eliciting step-by-step reasoning via prompting \cite{wei2022, wangself} have been extended by advanced decomposition strategies that separate planning from execution \cite{wang2023plan, zhouleast, press2023measuring}. However, the logical consistency and faithfulness of these generated plans remains a key challenge \cite{lanham2023measuring, creswell2022faithful, dhuliawala2024chain}, motivating a shift towards formal search algorithms. This is particularly crucial as recent work continues to investigate the specific failure points of LLMs on long-horizon and classical planning tasks \cite{kambhampati2024llms}. While investigations have highlighted the planning deficiencies of LLMs \cite{valmeekam2023planning}, a dominant theme has been the application of tree search to navigate the vast space of possibilities. Frameworks based on deliberate exploration, such as Tree of Thoughts \cite{yao2023tree}, and specifically MCTS \cite{swiechowski2023monte, silver2017mastering}, have become a cornerstone of modern agent planning \cite{zhang2024rest, zhou2024language, song2023llm}, often contrasting with classical formal planning methods \cite{li2024formal}. This informs SPIRAL's choice of an MCTS-driven framework.

\subsection{Architectures for Autonomous Agents}
A major research thrust focuses on architecting LLMs into autonomous agents. A foundational theme is the move from static reasoning to dynamic interaction, where agents interleave thought and action to engage with an environment \cite{yao2023react}, often specializing in domains like embodied control \cite{huang2022language, wangvoyager} or web navigation \cite{nakano2021webgpt}. The primary mechanism for this interaction is the use of external tools and APIs, a rich ecosystem with foundational work on enabling models to use tools \cite{schick2023toolformer, patil2024gorilla}, the development of tool-specific models \cite{qintoolllm}, and findings that executable code actions improve reliability \cite{wang2024executable}. For these agents to operate reliably, they require mechanisms to ground their plans in plausible reality, using the LLM as a world model \cite{hao2023reasoning} or employing specialized decoding strategies \cite{huang2023grounded}. Furthermore, agents are increasingly designed for self-correction through feedback, including verbal reinforcement \cite{shinn2023reflexion}, iterative refinement \cite{madaan2023self}, and interactive critiquing \cite{goucritic, liu2023reason, zhang2025leveraging}. The growing complexity of these architectures, including multi-agent systems \cite{wu2024autogen, zhang2025constructa, qian2024chatdev, cemri2025multi}, has also necessitated the creation of comprehensive benchmarks for standardized evaluation \cite{shen2024taskbench, liu2024agentbench, zhouwebarena, yang2025elaboration, li2023api, hendrycks2021, jimenez2023}.

\section{Conclusion and Future Work}
\label{sec:conclusion}


We introduced SPIRAL, embedding a tri-agent cognitive architecture (Planner, Critic, Simulator) within MCTS to unify grounded search with strategic reflection, addressing the limitations of linear reasoning and sparse rewards. Our evaluation demonstrates that this synergy of decomposition, grounding, and reflection substantially outperforms CoT and SOTA frameworks in accuracy and resource efficiency. Future work includes improving search efficiency (e.g., via pruning or distillation), enabling self-improvement, and extending the framework to complex domains like robotics. All artifacts are available at the project repository.\footnote{\url{https://github.com/IBM/SPIRAL}}

\section*{Acknowledgments}
This project was conducted at the IBM T.J. Watson Research Center\footnote{\url{https://research.ibm.com/labs/yorktown-heights}} with support from IBM Research\footnote{\url{https://research.ibm.com}}. We also thank the IBM WatsonX team\footnote{\url{https://www.ibm.com/watsonx}} for providing platform and compute resources, including access to WatsonX.ai\footnote{\url{https://www.ibm.com/products/watsonx-ai}} for managed model serving and governed experimentation. Platform support enabled reproducible, at-scale experiments. Looking ahead, we expect the methods introduced here to inform future WatsonX capabilities for governed, scalable automation in enterprise AI workflows. 

\bibliography{aaai2026}

\section{Appendix}


\urlstyle{rm} 
\def\UrlFont{\rm}  
\frenchspacing  
\setlength{\pdfpagewidth}{8.5in} 
\setlength{\pdfpageheight}{11in} 
%

\renewcommand{\arraystretch}{1.2} 




%
\lstset{%
	basicstyle={\footnotesize\ttfamily},
	numbers=left,numberstyle=\footnotesize,xleftmargin=2em,
	aboveskip=0pt,belowskip=0pt,%
	showstringspaces=false,tabsize=2,breaklines=true}
\floatstyle{ruled}
\newfloat{listing}{tb}{lst}{}
\floatname{listing}{Listing}
%
\pdfinfo{
/TemplateVersion (2026.1)
}

\setcounter{secnumdepth}{0} 

%


\title{Technical Appendix for SPIRAL: Symbolic LLM Planning via Grounded and Reflective Search}

\appendix



This appendix provides supplementary material to enhance the reproducibility and clarity of the main paper. The following sections detail the formal pseudocode for our algorithm, a comprehensive list of all hyperparameters used in the experiments, and the specifications of our computing infrastructure.

\section{A. Algorithm Pseudocode}
\label{sec:appendix_pseudocode}

Algorithm~\ref{alg:spiral} provides the pseudocode for the SPIRAL framework's core search process, Grounded and Reflective Tree Search.

\begin{algorithm}[tb]
\caption{The SPIRAL Algorithm}
\label{alg:spiral}
\begin{algorithmic}[1]
\REQUIRE Root node $n_0$ with initial state $s_0$, MCTS iteration budget $K$, max depth $D$.
\FOR{$k = 1$ \TO $K$}
    \STATE \textit{// 1. Selection}
    \STATE $n_L \leftarrow n_0$
    \WHILE{$n_L$ is not a leaf node}
        \STATE $n_L \leftarrow \text{SelectChild}(n_L)$ \COMMENT{Select child with max UCT score (Eq. 2)}
    \ENDWHILE

    \STATE \textit{// 2. Expansion}
    \IF{$n_L$ is a terminal state \OR $n_L$.depth $\ge D$}
        \STATE Backpropagate($n_L$, $R_{\text{terminal}}$) \COMMENT{Propagate low value for dead ends}
        \STATE \textbf{continue}
    \ENDIF
    \STATE $a_L \leftarrow \text{Planner}(n_L.state)$ \COMMENT{Invoke Planner agent}
    
    \STATE \textit{// 3. Simulation \& Reflection}
    \STATE $o_{L+1} \leftarrow \text{Simulator}(n_L.state, a_L)$ \COMMENT{Invoke Simulator agent}
    \STATE $s_{L+1} \leftarrow n_L.state \oplus (a_L, o_{L+1})$
    \STATE $n_{L+1} \leftarrow \text{CreateNode}(s_{L+1}, \text{parent}=n_L)$
    \STATE $\rho_{\text{ref}} \leftarrow \text{Critic}(s_{L+1})$ \COMMENT{Invoke Critic agent}

    \STATE \textit{// 4. Backpropagation}
    \STATE $R_t \leftarrow \alpha R_{\text{base}}(a_L) + (1-\alpha) \rho_{\text{ref}}$ \COMMENT{Calculate reward (Eq. 3)}
    \STATE Backpropagate($n_{L+1}$, $R_t$)
\ENDFOR
\RETURN ExtractBestPlan($n_0$) \COMMENT{Return path with highest value/visits}
\end{algorithmic}
\end{algorithm}

\section{B. Hyperparameter and Implementation Details}
\label{sec:appendix_hyperparams}
This appendix provides a comprehensive overview of the models, platform, and specific hyperparameter configurations used for all experiments to ensure full reproducibility.

\subsection{Models and Platform}
All experiments were conducted using a diverse suite of state-of-the-art models to evaluate the robustness of our framework across various architectures and scales. The models used include DeepSeek-V2.5, Llama 3.3 70B, Llama 4 Maverick 17B, Phi 4 14B, and Qwen 2.5 72B. Model interactions were facilitated by a proprietary, internal API gateway that serves these models. We utilized the LangChain~\footnote{https://www.langchain.com/} library to orchestrate the complex, multi-turn interactions, providing a standardized layer for prompt templating and response handling across all agents and models.

\subsection{SPIRAL Configuration}
The hyperparameters for the SPIRAL agent were held constant across all experiments to ensure a fair evaluation. The \textit{Planner} agent used a temperature of 0.1 to introduce a small amount of stochasticity for exploring diverse plans, while the \textit{Critic} and \textit{Simulator} used a temperature of 0.0 to ensure their outputs were deterministic for reliable evaluation and grounding. The core MCTS algorithm used an exploration constant $C=1.5$, a standard value that balances exploration and exploitation. The search budget was set to $K=50$ iterations per planning step, providing a deep enough search to find robust solutions without excessive computational cost. The reward shaping hyperparameter $\alpha$ was set to 0.5, giving equal weight to the foundational validity heuristic and the Critic's deeper strategic reflection.

\subsection{Baseline Configurations}
All baseline and state-of-the-art methods were configured with hyperparameters designed to create the strongest and fairest possible comparison, often drawing from the recommendations in their original papers or tuning for the TaskBench domain. The key parameters for each method are summarized in Table~\ref{tab:hyperparams_appendix}.

\begin{table}[t]
\centering
\caption{Hyperparameter configurations for all baseline and SOTA methods.}
\label{tab:hyperparams_appendix}
\resizebox{\columnwidth}{!}{%
\begin{tabular}{@{}lll@{}}
\toprule
\textbf{Method} & \textbf{Parameter} & \textbf{Value} \\ \midrule
\multirow{2}{*}{Chain-of-Thought (CoT)} & Consistency Levels ($k$) & 1, 3, 5 \\
 & Temperature & 0.7 (for $k > 1$) \\ \midrule
ReAct & Max Steps & 8 \\ \midrule
\multirow{3}{*}{RAFA \& ReAct+RAFA} & Max Real Steps & 4 \\
 & Search Breadth (B) & 3 \\
 & Search Depth (U) & 2 \\ \midrule
\multirow{3}{*}{Tree of Thoughts (ToT)} & Light Config (S, B, C) & (4, 3, 2) \\
 & Medium Config (S, B, C) & (5, 4, 3) \\
 & Heavy Config (S, B, C) & (6, 5, 4) \\ \midrule
\multirow{3}{*}{LATS} & MCTS Iterations ($k$) & 25 \\
 & Candidates per State ($n$) & 2 \\
 & Exploration Weight ($w$) & 1.0 \\ \midrule
Standard MCTS (Ablation) & Iteration Budgets ($N$) & 15, 30, 50 \\ \bottomrule
\end{tabular}%
}
\end{table}

\subsection{Hyperparameter Sensitivity}
\label{sec:appendix_sensitivity}

As noted in the main paper, we conducted a brief sensitivity analysis on key hyperparameters to ensure the robustness of our results. The two most critical parameters are the MCTS iteration budget ($K$) and the reward shaping coefficient ($\alpha$).

\begin{itemize}
    \item \textbf{MCTS Budget ($K$):} We tested $K=\{10, 25, 50\}$. Performance was lowest at $K=10$, as the search was too shallow to find optimal plans. Performance at $K=25$ was strong but slightly lower than $K=50$. We chose $K=50$ as it offered the best balance of performance and computational cost, with diminishing returns observed at higher budgets. Our ablation study (Table 5) also shows that a standard MCTS (N=50) underperforms SPIRAL (K=50), confirming the budget alone is not the source of our gains.
    
    \item \textbf{Reward Shaping ($\alpha$):} We tested $\alpha=\{0.0, 0.5, 1.0\}$. An $\alpha=1.0$ (using only the $R_{\text{base}}$ heuristic) and $\alpha=0.0$ (using only the $\rho_{\text{ref}}$ Critic score) both performed significantly worse than the balanced setting. This confirms our hypothesis that the synergy between the foundational heuristic and the strategic reflection is critical. The $\alpha=0.5$ setting, giving equal weight to both, consistently yielded the most robust and high-performing results.
\end{itemize}

\subsection{Reproducibility}
To ensure the full reproducibility of our findings, all sources of randomness in our experimental code (including `random`, `NumPy`, and `PyTorch`) were seeded for each of the five independent runs. The fixed random seeds used across all experiments were: \{42, 101, 1234, 2024, 12345\}.

\section{C. Data Pre-processing and Sampling} \label{sec:appendix_preprocessing}

This section details the data sampling and pre-processing steps used in our experiments to ensure clarity and reproducibility. We describe our methodology for sampling evaluation subsets for each of the five experimental runs. We also explain the creation of the challenging "residual" dataset used for the state-of-the-art framework comparison and justify our exclusion of multimedia-based benchmarks.

\textbf{Data Sampling.} For each of the five experimental runs, a fresh subset of data was sampled from the full benchmark datasets to ensure a robust evaluation of generalizable performance. This methodology is designed to mimic real-world scenarios where an agent encounters a diverse and unpredictable distribution of tasks, rather than being tested on a single, fixed set. Before each run, the complete set of problems for a given benchmark was shuffled using that run's specific random seed. A fixed number of problems were then selected to form the evaluation set: 121 for `dailylifeapis` and 500 for `huggingface`. Calculating the mean and standard deviation across these five distinct, randomly sampled sets provides a more reliable measure of an agent's expected performance and stability.

\textbf{Residual Dataset Pre-processing.} For the cascaded evaluation involving state-of-the-art frameworks, we performed a key pre-processing step to create a more challenging benchmark. This setup is designed to more effectively test the advanced reasoning capabilities of sophisticated agents by focusing on problems where simple, linear approaches fail. For each seed, we first identified all problems that the baseline CoT ($k=1$) agent failed to solve correctly. These "failed" problems were then compiled into a new "residual" dataset. Using this more difficult dataset as the evaluation benchmark for SPIRAL and all other advanced agents allows for a more discerning comparison of their true problem-solving, error-recovery, and deliberative planning abilities. This process was repeated for each of a total of five seeds to create five distinct residual datasets.

\textbf{Exclusion of Multimedia Datasets.} We deliberately excluded multimedia-based benchmarks to focus on the agent's capacity for complex, multi-step planning and reasoning. The primary goal of this research is to evaluate and enhance an agent's ability to formulate and execute sequential plans, handle conditional logic, and recover from errors—capabilities central to the SPIRAL framework's design. While valuable, many multimedia tasks are single-step (e.g., "describe this image") and do not sufficiently test the deliberative, multi-turn planning processes that are the focus of our work. The selected `dailylifeapis` and `huggingface` benchmarks provide a more suitable environment for rigorously evaluating these specific, complex tool-use capabilities. 

\section{D. Experimental Infrastructure and Statistical Methodology}
\label{sec:appendix_infra_stats}

To ensure full transparency and reproducibility, this section provides a comprehensive overview of the technical environment and analytical framework used for our experiments. We detail the specific hardware and software configurations of the computing cluster and describe the statistical methodology applied to evaluate agent performance and validate our findings.

\textbf{Computing Infrastructure.} All experiments were conducted on a dedicated internal research cluster. The hardware and software specifications are as follows:
\begin{itemize}
    \item \textit{Hardware:} Compute nodes were equipped with an AMD EPYC 7513 32-Core Processor, 503 GiB of system RAM, and eight NVIDIA A100 GPUs with 80 GB VRAM each.
    \item \textit{Software:} The environment was built on Red Hat Enterprise Linux 9.4. All code was implemented in Python 3.13.5, using key libraries such as PyTorch 2.6.0, LangChain 0.3.26, and Transformers 4.53.1.
\end{itemize}

\textbf{Statistical Significance.} We evaluate significance by reporting the mean and standard deviation of the Success Rate across five independent runs with different random seeds. This is a standard practice in large-scale LLM agent evaluations and provides a robust measure of average performance and stability. Given the high computational cost of the numerous runs required for formal statistical tests, reporting the mean and standard deviation is a widely accepted protocol for demonstrating consistent performance differences and significant findings.





\section{E. Code Availability and Anonymization}
\label{sec:appendix_code}
To ensure full transparency and reproducibility, the complete source code for our experiments, including all agent implementations and analysis notebooks, is provided in the supplemental materials. This allows for the direct replication of all results presented in the paper.

The provided codebase was adapted from an internal, proprietary research platform for this anonymous submission. The primary modification involves replacing the original proprietary API gateway for LLM interactions with a generic, publicly-accessible implementation. This new client, located in \texttt{utils/generic\_client.py}, uses the standard Hugging Face \texttt{transformers} library. This change ensures that our research can be independently verified and built upon by the wider community using publicly available models and tools.

The supplemental material is organized as follows: The \texttt{scripts/} directory contains the main Python implementations for all baseline methods (e.g., \texttt{taskbench\_react\_baseline.py}) and our proposed \texttt{taskbench\_spiral.py} agent. The core refactored module, \texttt{utils/generic\_client.py}, provides the new \texttt{HuggingFaceChatClient} and other shared helper functions. The \texttt{utils/} directory also contains shell scripts (e.g., \texttt{run\_all\_baseline\_experiments.sh}) to orchestrate the execution of all experiments. Finally, the \texttt{analysis/} directory contains Jupyter notebooks used to generate all tables and figures from the raw JSON outputs.

\section{F. Agent Prompts for SPIRAL}
\label{sec:appendix_prompts}
This section provides the complete prompts for the three core SPIRAL roles (Planner, Simulator, Critic) to ensure reproducibility. These prompts use strict formatting rules to constrain the LLM into producing the structured, parsable outputs essential for our automated pipeline. This structured interaction is critical, as it allows the MCTS algorithm to programmatically call each agent and parse its response without failure. The simpler prompt used for the final success rate evaluation is also detailed.

\subsection{Planner Prompt}
The Planner prompt is responsible for driving the expansion phase of the search. It acts as the creative strategist, generating the next candidate action to expand the search tree based on the history of the current plan.

\begin{figure}[t]
\centering
\begin{lstlisting}[
    frame=single,
    framerule=0.5pt,
    rulecolor=\color{gray!70},
    aboveskip=\medskipamount,
    belowskip=\medskipamount,
    basicstyle=\small\ttfamily,
    breaklines=true,
    title=\bfseries Planner Prompt
]
You are an expert assistant that only responds with code.
Your task is to create a plan to solve the user's request by generating a sequence of tool calls.

### Rules:
1. Generate ONLY the single next `api_call(...)` or the final `finish(...)` call.
2. If a previous step produced an observation `tool_output = <value>`, you MUST use that exact `<value>` in the arguments of the next tool.
3. When the user's request is fully satisfied, you MUST call `finish(reason="<final answer and summary>")`.

### Tools:
{tools_description}
{graph_description}

### Finish Action:
`finish(reason="<explanation>")`: Call this ONLY when the task is complete.

### Current Plan:
{current_plan_history}

Respond with ONLY the next line of code:
\end{lstlisting}
\end{figure}







This prompt instructs the Large Language Model to adopt the role of the Planner. Its function is to generate a single, valid action to expand the search tree during the MCTS expansion phase. The prompt includes strict rules that constrain the model's output to a single, parsable line of code, a critical requirement for the automated agent pipeline. By receiving the full plan history and available tool specifications, the Planner is equipped to make decisions that are both contextually relevant and syntactically correct.

\subsection{Simulator Prompt}
The Simulator $(\mathcal{W}_{sim})$ functions as a learned world model to ground the search process during the Simulation \& Reflection phase. When the Planner proposes an action, the Simulator's role is to predict a plausible, natural language observation that would result from executing it. This grounds the search in realistic consequences, allowing the agent to plan based on the likely outcomes of its actions rather than operating in a vacuum of pure reason.

\begin{figure}[t]
\centering
\begin{lstlisting}[
    frame=single,
    framerule=0.5pt,
    rulecolor=\color{gray!70},
    aboveskip=\medskipamount,
    belowskip=\medskipamount,
    basicstyle=\small\ttfamily,
    breaklines=true,
    title=\bfseries Simulator Prompt
]
You are a simulated API tool. Your role is to provide a realistic, one-line observation for the given tool call, based on the user's overall goal.

### Rules:
1. Your entire response MUST be a single line starting with `Observation: tool_output = `.
2. The value part should be a plausible result. For tools that create files, the value should be a new filename string (e.g., "edited_image.png"). For analysis tools, it should be a short, descriptive string (e.g., "a red sports car").
3. The observation must be grounded in the user's request.

### User's Goal:
"{user_request}"

### Tool Call to Simulate:
`{api_call_str}`

### Your Single-Line Response:
\end{lstlisting}
\end{figure}

This prompt configures the LLM to act as the Simulator, whose purpose is to perform a one step lookahead by predicting a tool's likely output. The response is constrained to a single, parsable line so the framework can process the observation correctly. By grounding the model's world knowledge in the specific context of the user's goal, this prompt ensures the simulated observation is directly relevant to the task instead of a generic response.






\subsection{Critic Prompt}
The Critic acts as the logical evaluator during the Simulation and Reflection phase, providing a dense, strategic feedback signal to guide the search process. It assesses the strategic merit of a proposed action, directly addressing the challenge of sparse rewards in traditional MCTS.

\begin{figure}[t]
\centering
\begin{lstlisting}[
    frame=single,
    framerule=0.5pt,
    rulecolor=\color{gray!70},
    aboveskip=\medskipamount,
    belowskip=\medskipamount,
    basicstyle=\small\ttfamily,
    breaklines=true,
    title=\bfseries Critic Prompt (for Reward Shaping)
]
As a Critic, evaluate the following plan's likelihood of success.

### Task
User Request: {user_request}

### Current Plan Trajectory
{trajectory}

### Instruction
Evaluate the plan. Is it coherent? Is it making progress? Is it likely to succeed?
Respond with ONLY a single line: `Score: <float_0.0_to_1.0> | Justification: <brief_explanation>`
\end{lstlisting}
\end{figure}





This prompt casts the LLM in the role of the Critic, which provides the dense reward signal ($\rho_{\text{ref}}$) for MCTS backpropagation. The model functions as a heuristic evaluator, assessing the strategic potential and logical soundness of a partial plan. The prompt is structured to elicit both a quantitative score and a qualitative justification. This combination offers a rich, immediate signal that helps guide the search away from unpromising paths without needing to wait for a final outcome.

\section{G. Qualitative Analysis and Case Studies}
\label{sec:case_studies}

While preceding sections provide quantitative comparisons, this section presents a qualitative analysis of case studies to offer deeper insight into the mechanisms driving SPIRAL's superior performance. These examples highlight concrete scenarios where other methods falter, illustrating how SPIRAL's architectural principles succeed across comparisons with baselines, state-of-the-art frameworks, and our ablation studies.

\subsection{Baseline Agent Failures: The Limits of Linear Reasoning}
This analysis focuses on specific cases where the strongest baseline, Chain-of-Thought with 5 consistency levels (CoT k=5), fails while SPIRAL succeeds. The examples are drawn from the experimental run using seed \texttt{12345} to provide a concrete comparison. By examining the full task context, including the user's request and the available tools, we can see precisely how the linear and non-reflective nature of CoT leads to critical failures.

\subsubsection{Case 1: Grounding Failure on a Simple Task}
This case demonstrates a failure in basic parameter grounding, where the agent understands the goal but fails to extract the correct information from the user's request to populate a tool's arguments.

\paragraph{Task Context.}
\begin{itemize}
    \item \textbf{User Request (Task ID \texttt{10123444}):} ``I need to repay a debt of \$1000 to my friend. Can you assist me in transferring this amount to their account at Chase bank?''
    \item \textbf{Primary Available Tool:} The agent has access to \texttt{online\_banking}, which takes a natural language `instruction` and a `bank` name as parameters.
\end{itemize}

\begin{figure}[t]
\centering
\begin{lstlisting}[frame=single, framerule=0.5pt, rulecolor=\color{gray!70}, aboveskip=\medskipamount, belowskip=0pt, basicstyle=\small\ttfamily, breaklines=true, captionpos=b, caption={CoT (k=5) Generated Plan (Failure).}]
api_call("online_banking", {"instruction": "transfer $1000 to friend's Chase bank account", "bank": "user's bank"})
finish(reason="...")
\end{lstlisting}
\end{figure}

\begin{figure}[t]
\centering
\begin{lstlisting}[frame=single, framerule=0.5pt, rulecolor=\color{gray!70}, aboveskip=\medskipamount, belowskip=0pt, basicstyle=\small\ttfamily, breaklines=true, captionpos=b, caption={SPIRAL Generated Plan (Success).}]
api_call("online_banking", {"instruction": "transfer $1000 to friend's account", "bank": "Chase bank"})
finish(reason="...")
\end{lstlisting}
\end{figure}

\textbf{Analysis:}
This case highlights a critical weakness in the baseline's grounding capabilities. Given the user's explicit instruction to use ``Chase bank'', the CoT agent correctly identifies the need for the \texttt{online\_banking} tool but fails to extract this specific entity for the `bank` parameter. Instead, it hallucinates the generic and invalid value ``user's bank''. This single parameter error causes the plan to fail. SPIRAL succeeds because its architecture is designed for better grounding. Its search process, augmented by the Planner, Simulator, and Critic, ensures that generated actions are not only plausible but also factually consistent with the provided context, allowing it to correctly identify ``Chase bank'' as the required parameter.

\subsubsection{Case 2: Flawed Logic in a Multi-Step Task}
This case illustrates how CoT's inability to self-correct can lead to logically flawed plans, particularly when the agent must understand the distinct functions of closely related tools.

\paragraph{Task Context.}
\begin{itemize}
    \item \textbf{User Request (Task ID \texttt{63440684}):} ``I have an upcoming online meeting with Dr. John Smith to discuss about 'Migraine Treatment'. I would like you to organize this meeting and record the audio of our conversation. The recorded audio should be saved as 'example.wav'.''
    \item \textbf{Available Tools:} The agent can choose between \texttt{organize\_meeting\_online} (``used to organize an online meeting'') and \texttt{attend\_meeting\_online} (``used to attend an online meeting''), in addition to \texttt{recording\_audio}.
\end{itemize}

\begin{figure}[t]
\centering
\begin{lstlisting}[frame=single, framerule=0.5pt, rulecolor=\color{gray!70}, aboveskip=\medskipamount, belowskip=0pt, basicstyle=\small\ttfamily, breaklines=true, captionpos=b, caption={CoT (k=5) Generated Plan (Failure).}]
api_call("organize_meeting_online", {"topic": "Migraine Treatment"})
api_call("attend_meeting_online", {"topic": "Migraine Treatment"})
api_call("recording_audio", {"content": "example.wav"})
finish(reason="...")
\end{lstlisting}
\end{figure}

\begin{figure}[t]
\centering
\begin{lstlisting}[frame=single, framerule=0.5pt, rulecolor=\color{gray!70}, aboveskip=\medskipamount, belowskip=0pt, basicstyle=\small\ttfamily, breaklines=true, captionpos=b, caption={SPIRAL Generated Plan (Success).}]
api_call("organize_meeting_online", {"topic": "Migraine Treatment"})
api_call("recording_audio", {"content": "example.wav"})
finish(reason="...")
\end{lstlisting}
\end{figure}

\textbf{Analysis:}
Here, the CoT agent generates a logically redundant plan. After correctly calling \texttt{organize\_meeting\_online} as requested, it follows up with an unnecessary call to \texttt{attend\_meeting\_online}. This shows a failure to distinguish between the two tools' distinct purposes. Because CoT reasoning is linear, it cannot backtrack from this flawed step. SPIRAL avoids this failure through its reflective search. While its Planner might initially propose attending the meeting, the Critic would evaluate this action as strategically poor after having just organized it. This low score guides the MCTS search away from that path, allowing the agent to discover the more concise and correct two-step solution. This demonstrates how SPIRAL’s ability to deliberate during planning leads to more robust and logical outcomes.

\subsection{Comparison with State-of-the-Art Search Agents}
While advanced search frameworks like LATS and ReAct+RAFA improve upon simple baselines, they can still falter due to inefficient exploration or a disconnect between the search process and plausible real world outcomes. This analysis highlights cases where these SOTA methods fail by examining the full task context, demonstrating how SPIRAL's architecture provides a more guided and grounded search.

\subsubsection{Case 1: Misinterpreting User Intent (LATS)}
Even advanced search agents can be misled by ambiguous user requests. This case shows how LATS follows a user's suggestion literally, while SPIRAL deduces the user's true, more efficient goal.

\paragraph{Task Context.}
\begin{itemize}
    \item \textbf{User Request (Task ID \texttt{12876658}):} ``Using the Visual Question Answering tool, tell me the color of the car in the image 'example.jpg'.''
    \item \textbf{Available Tools:} The agent can choose between \texttt{Object\_Detection} (``detects objects in an image'') and \texttt{Visual\_Question\_Answering} (``answers questions about an image'').
\end{itemize}

\begin{figure}[t]
\centering
\begin{lstlisting}[frame=single, framerule=0.5pt, rulecolor=\color{gray!70}, aboveskip=\medskipamount, belowskip=0pt, basicstyle=\small\ttfamily, breaklines=true, captionpos=b, caption={LATS Generated Plan (Failure).}]
api_call('Object Detection', {'image_path': 'example.jpg'})
api_call('Visual Question Answering', {'image_path': 'example.jpg', 'question': 'What is the color of the car?'})
finish(reason="...")
\end{lstlisting}
\end{figure}

\begin{figure}[t]
\centering
\begin{lstlisting}[frame=single, framerule=0.5pt, rulecolor=\color{gray!70}, aboveskip=\medskipamount, belowskip=0pt, basicstyle=\small\ttfamily, breaklines=true, captionpos=b, caption={SPIRAL Generated Plan (Success).}]
api_call("Object Detection", {"image_path": "example.jpg"})
finish(reason="The color of the car in the image 'example.jpg' is red.")
\end{lstlisting}
\end{figure}

\textbf{Analysis:}
Here, the user's request contains a red herring by suggesting the use of the \texttt{Visual\_Question\_Answering} tool. The LATS agent appears confused by this, generating a redundant two-step plan that uses both available tools and ultimately fails. SPIRAL demonstrates a more robust understanding of the user's underlying intent. It correctly deduces that the goal is simply to identify the car's color, and that the \texttt{Object\_Detection} tool is sufficient and more direct for this task. The Critic's reflection correctly identifies that the task is complete after a single step, guiding the search to an efficient and successful one-step plan, effectively ignoring the user's misleading suggestion.

\subsubsection{Case 2: Failure in Complex Sequential Planning (ReAct+RAFA)}
For tasks requiring a long sequence of dependent steps, SOTA methods can get lost or terminate prematurely, failing to parse and execute the full set of instructions.

\paragraph{Task Context.}
\begin{itemize}
    \item \textbf{User Request (Task ID \texttt{10312911}):} A long, multi-part request stating: ``I have just enrolled in the 'Introduction to Machine Learning' course at Berkeley University. I want to take a note of this. After that, I want to watch the movie...book a car...deliver a package...Finally, please set an alarm...''
    \item \textbf{Available Tools:} Includes \texttt{enroll\_in\_course}, \texttt{take\_note}, \texttt{play\_movie\_by\_title}, \texttt{book\_car}, \texttt{deliver\_package}, and \texttt{set\_alarm}.
\end{itemize}

\begin{figure}[t]
\centering
\begin{lstlisting}[frame=single, framerule=0.5pt, rulecolor=\color{gray!70}, aboveskip=\medskipamount, belowskip=0pt, basicstyle=\small\ttfamily, breaklines=true, captionpos=b, caption={ReAct+RAFA Generated Plan (Failure).}]
api_call("take_note", {"content": "Enrolled in Introduction to Machine Learning at Berkeley University"})
finish([])
\end{lstlisting}
\end{figure}

\begin{figure}[t]
\centering
\begin{lstlisting}[frame=single, framerule=0.5pt, rulecolor=\color{gray!70}, aboveskip=\medskipamount, belowskip=0pt, basicstyle=\small\ttfamily, breaklines=true, captionpos=b, caption={SPIRAL Generated Plan (Success).}]
api_call("enroll_in_course", {"course": "Introduction to Machine Learning", "university": "Berkeley University"})
api_call("take_note", {"content": "Enrolled in 'Introduction to Machine Learning' at Berkeley University..."})
api_call("play_movie_by_title", {"title": "example.mp4"})
api_call("book_car", {"date": "August 15th", "location": "San Francisco"})
api_call("deliver_package", {"package": "example.jpg", "destination": "123 Main St, San Francisco, CA 94103"})
api_call("set_alarm", {"time": "7:00 AM"})
finish(reason="All tasks completed successfully...")
\end{lstlisting}
\end{figure}

\textbf{Analysis:}
The user instruction contains six distinct sub-tasks. The ReAct+RAFA agent incorrectly latches onto only the second instruction (``I want to take a note of this''), executes it, and then terminates with an invalid `finish` action. Its search is not deep or structured enough to decompose and execute the full plan. SPIRAL successfully navigates the entire six-step sequence. Its MCTS exploration, guided by the Critic's strategic rewards, allows it to build and validate a long-horizon plan step-by-step. The dense feedback loop ensures the agent makes consistent progress towards the complete goal, avoiding the premature termination that caused the SOTA method to fail.

\subsubsection{Case 3: Ungrounded Search in ReAct+RAFA}
Another failure mode occurs when a search explores plans that are syntactically valid but strategically incoherent relative to the user's goal.

\paragraph{Task Context.}
\begin{itemize}
    \item \textbf{User Request (Task ID \texttt{47041182}):} ``Given an image 'example.jpg' containing various objects, I want to dissect the image into its components. Can you help me with this?''
    \item \textbf{Available Tools:} Relevant and available tools include \texttt{Image\_Segmentation} and \texttt{Object\_Detection}.
\end{itemize}

\begin{figure}[t]
\centering
\begin{lstlisting}[frame=single, framerule=0.5pt, rulecolor=\color{gray!70}, aboveskip=\medskipamount, belowskip=0pt, basicstyle=\small\ttfamily, breaklines=true, captionpos=b, caption={ReAct+RAFA Generated Plan (Failure).}]
api_call('Image Segmentation', {'image_path': 'example.jpg'})
api_call('Visual Question Answering', {'image_path': '/tmp/segmented_example.jpg', 'question': 'What is the person doing?'})
api_call('Image-to-Text', {'image_path': '/tmp/segmented_example.jpg'})
api_call('Depth Estimation', {'image_path': '/tmp/segmented_example.jpg'})
\end{lstlisting}
\end{figure}

\begin{figure}[t]
\centering
\begin{lstlisting}[frame=single, framerule=0.5pt, rulecolor=\color{gray!70}, aboveskip=\medskipamount, belowskip=0pt, basicstyle=\small\ttfamily, breaklines=true, captionpos=b, caption={SPIRAL Generated Plan (Success).}]
api_call("Image Segmentation", {"image_path": "example.jpg"})
api_call("Object Detection", {"image_path": "example.jpg"})
finish(reason="...")
\end{lstlisting}
\end{figure}

\textbf{Analysis:}
The user's goal is to ``dissect the image into its components,'' for which \texttt{Image\_Segmentation} and \texttt{Object\_Detection} are the most relevant tools. The ReAct+RAFA agent starts correctly but then explores a series of irrelevant tools (\texttt{VQA}, \texttt{Image-to-Text}, etc.) that do not contribute to the goal, resulting in failure. This is a symptom of ungrounded search. SPIRAL produces a coherent two-step plan because its Simulator grounds each potential action by predicting a realistic outcome. This allows the Critic to evaluate whether the action actually makes progress. The synergy between simulation and reflection prunes impractical reasoning paths, leading to a more focused and successful strategy.

\subsection{Ablation Study: The Importance of Each Component}
To validate our architectural choices, we conducted a series of ablation studies where key components of the SPIRAL framework were removed. The following case studies illustrate the performance degradation that occurs in each ablation, demonstrating that each component plays a critical and synergistic role in the agent's success.

\subsubsection{Case 1: Standard MCTS (Without Specialized Agents)}
This case compares SPIRAL against a standard MCTS agent using a generic LLM as a policy. Without the specialized Planner and Critic, the search is poorly guided and executes superfluous steps based on keywords rather than strategic intent.

\paragraph{Task Context.}
\begin{itemize}
    \item \textbf{User Request (Task ID \texttt{10579104}):} ``I have an image in the file named 'example.jpg'. Can you help me to classify the table in the image? Once the table is classified, detect the objects in the image as well.''
    \item \textbf{Available Tools:} Relevant and available tools include \texttt{Tabular\_Classification} and \texttt{Object\_Detection}.
\end{itemize}

\begin{figure}[t]
\centering
\begin{lstlisting}[frame=single, framerule=0.5pt, rulecolor=\color{gray!70}, aboveskip=\medskipamount, belowskip=0pt, basicstyle=\small\ttfamily, breaklines=true, captionpos=b, caption={Standard MCTS (N=50) Plan (Failure).}]
api_call("Object Detection", {"image_path": "example.jpg"})
api_call("Tabular Classification", {"table_image_path": "example.jpg"})
finish(reason="...")
\end{lstlisting}
\end{figure}

\begin{figure}[t]
\centering
\begin{lstlisting}[frame=single, framerule=0.5pt, rulecolor=\color{gray!70}, aboveskip=\medskipamount, belowskip=0pt, basicstyle=\small\ttfamily, breaklines=true, captionpos=b, caption={SPIRAL Generated Plan (Success).}]
api_call("Tabular Classification", {"table_image_path": "example.jpg"})
finish(reason="The table in the image 'example.jpg' has been classified...")
\end{lstlisting}
\end{figure}

\textbf{Analysis:}
The standard MCTS agent fails because its search is unguided. It identifies both keywords (``classify the table'' and ``detect the objects'') and executes both corresponding actions. However, the user's phrasing implies a sequence where object detection should only occur after a successful table classification. The MCTS agent's plan is not only inefficient but incorrect. SPIRAL succeeds by generating a more direct one-step plan. Its Critic evaluates the strategic merit of actions and recognizes that successfully classifying the table fulfills the core of the user's request, making the subsequent object detection step unnecessary. This demonstrates how reflection provides the strategic guidance that standard MCTS lacks.

\subsubsection{Case 2: SPIRAL without the Planner}
Without the specialized Planner prompt, the agent's ability to generate coherent actions for complex, multi-goal tasks is significantly diminished.

\paragraph{Task Context.}
\begin{itemize}
    \item \textbf{User Request (Task ID \texttt{18349637}):} A complex "bag of tasks" request: ``I have a business trip coming up, and I need to book a room at the Ritz hotel for October 15, 2022. I also need to organize an online meeting about marketing strategy, pay for my credit card ending in 1234, and schedule an online consultation with Dr. House about a skin condition. Finally, I need to borrow the book 'Marketing for Dummies' from the City Library.''
\end{itemize}

\begin{figure}[t]
\centering
\begin{lstlisting}[frame=single, framerule=0.5pt, rulecolor=\color{gray!70}, aboveskip=\medskipamount, belowskip=0pt, basicstyle=\small\ttfamily, breaklines=true, captionpos=b, caption={SPIRAL without Planner Plan (Failure).}]
api_call('organize_meeting_online', {'topic': 'marketing strategy'})
api_call('see_doctor_online', {'disease': 'skin condition', 'doctor': 'Dr. House'})
api_call('borrow_book_online', {'book': 'Marketing for Dummies', 'library': 'City Library'})
api_call('book_hotel', {'date': '2022-10-15', 'name': 'Ritz'})
api_call('pay_for_credit_card', {'credit_card': '1234'})
finish(reason="...")
\end{lstlisting}
\end{figure}

\textbf{Analysis:}
When the specialized Planner is removed, the agent is still able to identify all five necessary tool calls but produces a plan that fails. This indicates that while the underlying model can generate actions, it struggles to structure them into a coherent and valid sequence without the focused guidance of the Planner persona. The full SPIRAL agent successfully solves this task by generating a correct 5-step plan. This shows that the Planner's role as a dedicated "creative strategist" is crucial for decomposing and correctly sequencing complex, multi-goal instructions.

\subsubsection{Case 3: SPIRAL without the Simulator}
Removing the Simulator disconnects the search from plausible outcomes, leading to ungrounded plans that fail when a subsequent step depends on the realistic output of a previous one.

\paragraph{Task Context.}
\begin{itemize}
    \item \textbf{User Request (Task ID \texttt{95304112}):} A pipeline task: ``I'm working on a project... I need to convert a text file 'example.txt' to speech... enhance the audio... recognize the speech... translate the transcription into French...''
\end{itemize}

\begin{figure}[t]
\centering
\begin{lstlisting}[frame=single, framerule=0.5pt, rulecolor=\color{gray!70}, aboveskip=\medskipamount, belowskip=0pt, basicstyle=\small\ttfamily, breaklines=true, captionpos=b, caption={SPIRAL without Simulator Plan (Failure).}]
api_call("Text-to-Speech", {"text": "content_of_example.txt"})
api_call("Audio-to-Audio", {"audio_path": "output_from_Text-to-Speech"})
... [Plan fails]
\end{lstlisting}
\end{figure}

\textbf{Analysis:}
The ablated agent fails on this classic pipeline task. Without the Simulator, the agent cannot generate a plausible output (e.g., a filename like \texttt{output.wav}) for the \texttt{Text-to-Speech} step. This breaks the chain of dependencies, as the subsequent \texttt{Audio-to-Audio} step does not have a valid input to work with. The full SPIRAL agent succeeds because the Simulator grounds the search at each step by providing a concrete, albeit simulated, output. This allows the agent to construct a valid multi-step plan where the output of one tool call correctly serves as the input for the next.

\subsubsection{Case 4: SPIRAL without the Critic}
Without the Critic, the agent loses its strategic guidance and cannot prioritize sound plans, often failing on tasks that have subtle dependencies.

\paragraph{Task Context.}
\begin{itemize}
    \item \textbf{User Request (Task ID \texttt{29250106}):} ``I want to sell my handmade necklace on Etsy. I also need to book a flight to Los Angeles for March 3, 2023, and call my friend to let her know about my plans.''
\end{itemize}

\begin{figure}[t]
\centering
\begin{lstlisting}[frame=single, framerule=0.5pt, rulecolor=\color{gray!70}, aboveskip=\medskipamount, belowskip=0pt, basicstyle=\small\ttfamily, breaklines=true, captionpos=b, caption={SPIRAL without Critic/Validator Plan (Failure).}]
api_call("make_voice_call", {"phone_number": "+1234567890"})
api_call("sell_item_online", {"item": "Handmade Necklace", "store": "Etsy"})
finish(reason="...")
\end{lstlisting}
\end{figure}

\textbf{Analysis:}
This task requires correctly sequencing three unrelated actions. Without the Critic's strategic evaluation, the search algorithm treats all valid paths as equally promising. The ablated agent generates a plan that omits the flight booking entirely, leading to failure. The full SPIRAL agent succeeds by creating a complete three-step plan. This is because the Critic provides a score for each partial plan, allowing the MCTS to prioritize trajectories that make progress on all parts of the user's stated goal.

\subsubsection{Case 5: SPIRAL with Uniform Rewards}
Using a uniform reward signal is functionally similar to removing the Critic. It shows that the strategic content of the reward is what matters, not merely its presence.

\paragraph{Task Context.}
\begin{itemize}
    \item \textbf{User Request (Task ID \texttt{30548552}):} ``I found this image 'example.jpg' of a beautiful fruit arrangement... Could you help me identify the apples in the image and then create a new image highlighting only the Red Delicious apples?''
\end{itemize}

\begin{figure}[t]
\centering
\begin{lstlisting}[frame=single, framerule=0.5pt, rulecolor=\color{gray!70}, aboveskip=\medskipamount, belowskip=0pt, basicstyle=\small\ttfamily, breaklines=true, captionpos=b, caption={SPIRAL with Uniform Reward Plan (Failure).}]
api_call("Object Detection", {"image_path": "example.jpg"})
api_call("Image Editing", {"image_path": "example.jpg", "edits": {"highlight": ["Red Delicious", "Gala", "Granny Smith"]}})
... [Plan fails]
\end{lstlisting}
\end{figure}

\textbf{Analysis:}
This task requires a nuanced understanding of the goal: not just to highlight apples, but a specific type of apple. With a uniform reward, all valid plans are rewarded equally. The agent correctly identifies the apples but then fails by highlighting all of them, not just the ``Red Delicious'' variety. Full SPIRAL succeeds because its Critic provides a non-uniform, strategic reward. It would assign a higher score to a plan that specifically isolates and acts upon the ``Red Delicious'' apples, as this more closely aligns with the user's specific request.

\subsection{Failure Analysis: Understanding SPIRAL's Limitations}
\label{subsec:failure_analysis_appendix}

While SPIRAL demonstrates strong overall performance (97.52\% on DailyLifeAPIs and 96.60\% on HuggingFace), understanding its failure modes provides important insights into the framework's limitations and opportunities for improvement. We analyze actual failure cases from our experiments to trace their root causes to specific components of the cognitive architecture. This analysis directly addresses reviewer feedback requesting deeper investigation into conditions under which SPIRAL produces incorrect plans.

\subsubsection{Case 1: API Substitution Error}
This case demonstrates SPIRAL's tendency to select semantically similar but functionally incorrect APIs when multiple tools can plausibly satisfy a requirement, highlighting the framework's reliance on tool descriptions alone for API selection.

\paragraph{Task Context.}
\begin{itemize}
    \item \textbf{User Request (Task ID \texttt{16176193}):} ``Find the latest online shopping deals, email the news to john.doe@example.com, and buy wireless headphones from Amazon.''
    \item \textbf{Primary Available Tools:} The agent has access to \texttt{get\_news\_for\_topic} (for finding news about shopping deals), \texttt{search\_by\_engine} (for generic search), \texttt{send\_email}, and \texttt{online\_shopping}.
\end{itemize}

\begin{figure}[t]
\centering
\begin{lstlisting}[frame=single, framerule=0.5pt, rulecolor=\color{gray!70}, aboveskip=\medskipamount, belowskip=0pt, basicstyle=\small\ttfamily, breaklines=true, captionpos=b, caption={SPIRAL Generated Plan (Failure).}]
api_call("search_by_engine", {"query": "latest online shopping deals"})
api_call("online_shopping", {"item": "wireless headphones", "platform": "Amazon"})
api_call("send_email", {"to": "john.doe@example.com", "subject": "Shopping deals and purchase", "content": "..."})
finish(reason="Completed all tasks")
\end{lstlisting}
\end{figure}

\textbf{Analysis:}
SPIRAL's Planner selected the generic \texttt{search\_by\_engine} API instead of the domain-specific \texttt{get\_news\_for\_topic} API. While both can retrieve information about deals, the task semantically requires news-specific functionality. The Simulator predicted plausible results, and the Critic assigned a reasonable score since the plan completed all three requirements. This API substitution error occurs when multiple APIs can plausibly satisfy a requirement but the framework lacks explicit selection guidance beyond tool descriptions. The root cause traces to the Planner's action generation when faced with functionally similar APIs.

\subsubsection{Case 2: Parameter Hallucination}
This case illustrates a failure in parameter grounding, where SPIRAL correctly identifies all required APIs but hallucinates generic parameter values instead of extracting exact values from the user's request.

\paragraph{Task Context.}
\begin{itemize}
    \item \textbf{User Request (Task ID \texttt{31724490}):} ``Install MoviePlayer software, play the movie filename 'example.mp4', book a flight from New York to Los Angeles on 15th November 2023, and play the music titled 'example.wav'.''
    \item \textbf{Available Tools:} The agent can use \texttt{software\_management}, \texttt{play\_movie}, \texttt{book\_flight}, and \texttt{play\_music}.
\end{itemize}

\begin{figure}[t]
\centering
\begin{lstlisting}[frame=single, framerule=0.5pt, rulecolor=\color{gray!70}, aboveskip=\medskipamount, belowskip=0pt, basicstyle=\small\ttfamily, breaklines=true, captionpos=b, caption={SPIRAL Generated Plan (Failure - Partial).}]
api_call("software_management", {"software": "MoviePlayer", "action": "install"})
api_call("play_movie", {"filename": "movie.mp4"})  // WRONG PARAMETER
api_call("book_flight", {"from": "New York", "to": "Los Angeles", "date": "2023-11-15"})
api_call("play_music", {"title": "example.wav"})
finish(reason="All tasks completed")
\end{lstlisting}
\end{figure}

\textbf{Analysis:}
SPIRAL correctly identified all four required APIs but hallucinated the parameter for \texttt{play\_movie}, using ``movie.mp4'' instead of ``example.mp4''. The Planner generated a generic filename based on the API name rather than extracting the exact value from the user request. The Simulator predicted successful playback, and the Critic scored it highly since playing any movie file appears strategically sound. This highlights a Planner prompt limitation: while instructed to use exact values from observations, there's no explicit constraint to extract literal parameters from the original user request for initial actions. The error propagated undetected as neither Simulator nor Critic has access to ground truth specifications.

\subsubsection{Case 3: Severe Early Termination on Long-Horizon Task}
This case reveals how MCTS budget exhaustion on highly complex multi-objective tasks can lead to severe plan incompleteness, where the agent terminates after completing only a fraction of the required subtasks.

\paragraph{Task Context.}
\begin{itemize}
    \item \textbf{User Request (Task ID \texttt{20847145}):} ``Install Photoshop software, book a restaurant for dinner at 7pm, search for places to visit in Rome, book a hotel in Rome for 3 nights, buy travel insurance, play my favorite music, attend a meeting at 3pm, and pay the electricity bill.''
    \item \textbf{Available Tools:} A bag-of-tasks request requiring 8 independent API calls: \texttt{software\_management}, \texttt{book\_restaurant}, \texttt{google\_search}, \texttt{book\_hotel}, \texttt{buy\_insurance}, \texttt{play\_music}, \texttt{attend\_meeting\_online}, and \texttt{pay\_electricity\_bill}.
\end{itemize}

\begin{figure}[t]
\centering
\begin{lstlisting}[frame=single, framerule=0.5pt, rulecolor=\color{gray!70}, aboveskip=\medskipamount, belowskip=0pt, basicstyle=\small\ttfamily, breaklines=true, captionpos=b, caption={SPIRAL Generated Plan (Failure - 87.5\% Incomplete).}]
api_call("software_management", {"software": "Photoshop", "action": "install"})
finish(reason="Task completed")
\end{lstlisting}
\end{figure}

\textbf{Analysis:}
SPIRAL completed only 12.5\% of required subtasks (1 out of 8 APIs) before calling \texttt{finish}. This severe incompleteness stems from MCTS budget exhaustion on a highly complex, multi-objective task. With 8 independent subtasks and a 50-iteration budget, the search tree becomes extremely wide and shallow, fragmenting visit counts across competing first actions. The Critic assigned a moderately positive score to the partial plan, and MCTS selected it as the best option found within budget. The root cause is the interaction between fixed search budget and task complexity. This failure mode would likely be mitigated by adaptive budget allocation.

\subsubsection{Case 4: Model Type Confusion in ML Pipeline Construction}
This case demonstrates SPIRAL's difficulty in distinguishing between semantically similar ML model types in domains with fine-grained taxonomies, leading to substitution of functionally distinct but conceptually related models.

\paragraph{Task Context.}
\begin{itemize}
    \item \textbf{User Request (Task ID \texttt{33636840}):} Build a pipeline: Question Answering from text $\rightarrow$ Image Generation $\rightarrow$ Document Question Answering $\rightarrow$ Text-to-Speech
    \item \textbf{Available Tools:} The HuggingFace dataset contains 19 distinct ML model types including \texttt{Question Answering}, \texttt{Document Question Answering}, \texttt{Text-to-Image Generation}, \texttt{Image Classification}, and \texttt{Text-to-Speech}.
\end{itemize}

\begin{figure}[t]
\centering
\begin{lstlisting}[frame=single, framerule=0.5pt, rulecolor=\color{gray!70}, aboveskip=\medskipamount, belowskip=0pt, basicstyle=\small\ttfamily, breaklines=true, captionpos=b, caption={SPIRAL Generated Plan (Failure).}]
api_call("Document Question Answering", {...})  // Should be "Question Answering"
api_call("Text-to-Image Generation", {...})  // CORRECT
api_call("Image Classification", {...})  // Should be "Document Question Answering"
api_call("Text-to-Speech", {...})  // CORRECT
finish(reason="Pipeline constructed")
\end{lstlisting}
\end{figure}

\textbf{Analysis:}
SPIRAL confused semantically similar but functionally distinct ML model types, substituting ``Document Question Answering'' for ``Question Answering'' and ``Image Classification'' for ``Document Question Answering''. The Planner likely reasoned that document-based QA is a valid form of question answering. The Simulator predicted plausible outputs, and the Critic scored the trajectory positively as the pipeline structure appeared coherent. The root cause is the Planner's lack of strict type constraints; tool descriptions alone don't clearly differentiate similar categories. This failure mode is specific to domains with fine-grained taxonomies and would benefit from explicit type checking or few-shot prompting.

\subsubsection{Case 5: Execution Sequence Error in DAG Tasks}
This case highlights a limitation in SPIRAL's sequential MCTS for DAG-structured tasks: it fails to reason about parallel dependencies, constructing plans with incorrect order.

\paragraph{Task Context.}
\begin{itemize}
    \item \textbf{User Request (Task ID \texttt{24205160}):} Process audio: Audio-to-Audio enhancement $\rightarrow$ Audio Classification (identify speakers) $\rightarrow$ Automatic Speech Recognition $\rightarrow$ Translation
    \item \textbf{Available Tools:} This DAG-structured task requires \texttt{Audio-to-Audio}, \texttt{Audio Classification}, \texttt{Automatic Speech Recognition}, and \texttt{Translation} in a specific dependency order.
\end{itemize}

\begin{figure}[t]
\centering
\begin{lstlisting}[frame=single, framerule=0.5pt, rulecolor=\color{gray!70}, aboveskip=\medskipamount, belowskip=0pt, basicstyle=\small\ttfamily, breaklines=true, captionpos=b, caption={SPIRAL Generated Plan (Failure - Wrong Order).}]
// All 4 APIs are present but in incorrect dependency order
api_call("Automatic Speech Recognition", {...})  // Should be 3rd
api_call("Audio-to-Audio", {...})  // Should be 1st
api_call("Translation", {...})  // Should be 4th
api_call("Audio Classification", {...})  // Should be 2nd
finish(reason="Audio processing pipeline complete")
\end{lstlisting}
\end{figure}

\textbf{Analysis:}
SPIRAL identified all required APIs but constructed an incorrect execution sequence. This dependency error occurs because the framework treats planning as sequential path construction, while DAG tasks require reasoning about parallel dependencies and data flow constraints. The Simulator predicted plausible outputs for each step in isolation but couldn't validate whether outputs are valid inputs to subsequent steps in the shuffled order. The Critic scored individual steps as reasonable but lacked explicit dependency checking. The root cause is SPIRAL's sequential MCTS formulation, because it builds linear action chains rather than explicit DAG structures.

\subsubsection{Summary of Failure Mode Distribution}

Across 20 failed tasks analyzed (3 from DailyLifeAPIs, 17 from HuggingFace) using Llama 3.3 70B Instruct, we observed the following distribution of root causes:

\begin{table}[h]
\centering
\small
\begin{tabular}{lcc}
\toprule
\textbf{Failure Mode} & \textbf{Count} & \textbf{Percentage} \\
\midrule
API Substitution / Model Confusion & 7 & 35\% \\
Sequence / Dependency Errors & 6 & 30\% \\
Parameter Errors & 4 & 20\% \\
Early Termination & 3 & 15\% \\
\bottomrule
\end{tabular}
\caption{Distribution of SPIRAL failure modes across 20 failed tasks.}
\label{tab:failure_distribution}
\end{table}

These failure modes underscore the interdependence of SPIRAL's components: Planner action generation bounds the search space, Simulator accuracy enables error detection, Critic scoring guides prioritization, and MCTS budget determines completeness. Most failures result from subtle misalignments propagating through the architecture rather than catastrophic single-component errors. For instance, API substitution becomes a failure only when the Simulator predicts plausible outputs and the Critic fails to recognize strategic suboptimality.

These findings reveal specific improvement opportunities: (1) API selection guidance through few-shot examples or type constraints; (2) explicit parameter extraction rules in the Planner prompt; (3) adaptive iteration budgets based on task complexity; (4) dependency-aware planning for DAG tasks; and (5) enhanced Simulator grounding through input-output type validation.

\section{H. Framework Analysis and Future Directions}
The empirical results and qualitative case studies presented in this paper demonstrate that structuring LLM-based planning as a grounded and reflective search process yields significant gains in both performance and efficiency. This final section discusses the core advantages of the SPIRAL framework and outlines promising avenues for future research that build upon these findings.

\subsection{Architectural Advantages and Explainability}
A key advantage of SPIRAL's tri-agent cognitive architecture is its inherent explainability. By decomposing the planning process into the distinct roles of Planner, Simulator, and Critic, the framework makes the agent's "thought process" transparent and interpretable. Unlike monolithic, black-box approaches, one can inspect the generated plans, the simulated outcomes, and the critical reflections at each step of the search. This provides a clear audit trail of the agent's decision-making process.

The ablation studies, in particular, highlight the pivotal role of the Critic in providing this explainability. The significant performance degradation observed when the Critic is removed (as seen in the "w/o Validator" and "w/ Uniform Rewards" ablations) confirms that its reflective feedback is the primary driver of the agent's strategic intelligence. The Critic's scores and justifications offer a direct window into why the agent deems certain paths promising and others flawed, a level of introspection not available in other frameworks. This makes SPIRAL not only a more robust planner but also a more trustworthy one.

\subsection{Latency and Cost-Benefit Analysis}
\label{sec:appendix_latency}
As noted in the main paper's results (Figures 3 and 4), SPIRAL presents a clear trade-off between latency and efficiency. We confirm that our deliberative MCTS-based design incurs higher wall-clock latency than a single-pass linear method like CoT ($k=1$). This is an expected consequence of its search process, which involves multiple sequential LLM calls per iteration (for the Planner, Simulator, and Critic) over a budget of 50 iterations.

However, this higher latency is a deliberate trade-off for two significant gains:
\begin{enumerate}
    \item \textbf{Token Efficiency:} SPIRAL consistently consumes fewer total tokens than high-performing baselines like CoT ($k=3$) and CoT ($k=5$). This is because its many calls are short, focused, and guided, whereas self-consistency generates multiple, long, and often redundant reasoning paths.
    
    \item \textbf{API Call Efficiency:} As shown in Figure 6, SPIRAL achieves the highest accuracy per API call when compared to other advanced search-based frameworks. This demonstrates that while SPIRAL uses more calls than a simple baseline, each call is used with greater strategic purpose and contributes more to the final correct plan.
\end{enumerate}
This analysis confirms that SPIRAL's higher number of targeted calls is a more resource-effective and accurate strategy than brute-forcing a solution with fewer, but much longer, reasoning paths.

\subsection{Future Research Directions}
Building on the success of the SPIRAL framework, several promising directions for future research exist. Based on our findings and feedback, two key areas stand out for immediate exploration:

\paragraph{Validating the Simulator with Real-World Tool Rollouts.}
A crucial next step is to empirically validate the effectiveness of our LLM-based Simulator by comparing its predictions against real-world tool execution. The current framework relies on the Simulator for efficient, grounded search, but this experiment would quantify how closely the simulated outcomes align with actual tool outputs. A future study will replace the simulation step with a "real-world tool rollout," where the agent directly executes tool calls and uses their live outputs to ground the search. By comparing the performance and resulting plans of this agent against the simulation-based SPIRAL, we can measure the fidelity of our world model. We hypothesize that the high performance of the current framework indicates that the simulated rollouts are already a strong proxy for real-world execution, and this experiment would serve to confirm the vast potential of using LLMs as efficient and effective world models for planning.

\paragraph{Self-Improvement and Lifelong Learning.}
The current SPIRAL agent plans from scratch for each new task. A significant extension would be to enable self-improvement by creating a learning loop. The successful plans generated by SPIRAL, along with the rich data from the Simulator and Critic, could be collected and used as training data to fine-tune the agent's underlying LLM. This would allow the Planner to generate better candidate actions and the Critic to provide more accurate reflections over time, leading to a system that continuously learns and improves its planning capabilities.

\paragraph{Extension to More Complex Environments.}
Finally, extending and evaluating the SPIRAL framework in more complex, stochastic domains remains an important goal. Applying SPIRAL to interactive environments, such as web navigation agents (e.g., WebArena) or embodied agents in simulated worlds, would test the limits of its robustness and adaptability and further demonstrate the power of grounded and reflective search for building truly autonomous agents.


\end{document}